\documentclass{article} 
\usepackage[final]{colm2026_conference}

\usepackage{microtype}
\usepackage[most]{tcolorbox}
\usepackage{hyperref}

\usepackage[T1]{fontenc}
\usepackage{listings}
\usepackage{listingsutf8}
\usepackage{listings}
\usepackage{url}
\usepackage{booktabs}
\usepackage{amsmath}
\usepackage{caption} 
\usepackage{caption}
\usepackage{subcaption}
\usepackage{booktabs}
\usepackage{adjustbox}
\usepackage{multirow}
\usepackage{listings}
\usepackage{tcolorbox}
\tcbuselibrary{skins, breakable}
\usepackage{enumitem}
\usepackage{cleveref}
\usepackage[utf8]{inputenc}  
\usepackage{xcolor}
\tcbuselibrary{listings, breakable, skins}
\usepackage[most]{tcolorbox}
\usepackage{listings}
\usepackage{caption}
\usepackage{placeins}

\usepackage{lineno}

\definecolor{darkblue}{rgb}{0, 0, 0.5}
\hypersetup{colorlinks=true, citecolor=darkblue, linkcolor=darkblue, urlcolor=darkblue}

\DeclareMathSizes{10}{9}{6}{5}

\usepackage{colortbl}

\definecolor{berkeleyblue}{HTML}{3B7EA1}
\definecolor{berkeleygold}{HTML}{FDB515}
\newcommand\hy{\rowcolor{berkeleygold!40}}
\newcommand\hb{\rowcolor{berkeleygold!20}}

\newcommand{\OURS}[1]{INSIDE}
\title{\textsc{INSIDE} the Student's Mind: Jointly Modeling Latent Reasoning and Action in LLM Student Simulators}

\author{Rose Niousha\thanks{Corresponding author: \texttt{rose.n@berkeley.edu}}, \ Minwoo Kang, \ Narges Norouzi \\
Department of Electrical Engineering and Computer Sciences\\
University of California, Berkeley \\
\texttt{\{rose.n, minwoo\_kang, norouzi\}@berkeley.edu} \\
}

\begin{document}

\ifcolmsubmission
\linenumbers
\fi

\maketitle

\begin{abstract}
    Large Language Model (LLM)-based simulators often reproduce observable actions but fail to capture the underlying reasoning behind them. In education, where student simulation is increasingly used for various applications such as evaluating tutoring systems, this gap is especially pronounced. Two students may submit identical submissions for entirely different reasons. We present \textbf{\textsc{Internal Student Dialogue (INSIDE)}}, a student modeling framework that fine-tunes LLMs not only to act like students but also to think like them. INSIDE generates internal dialogue grounded in Bloom's Taxonomy across cognitive, affective, and action dimensions, and fine-tunes models on paired think traces and actions. We baseline against different prompting frameworks and evaluate on two axes: fidelity of simulated actions and quality of generated internal dialogue. Our evaluations show that INSIDE improves simulation fidelity in both action fidelity, matching code generation of real students, and reasoning alignment, achieving the highest alignment across models up to 57.9\%.

\end{abstract}
\section{Introduction}
\label{sec:introduction}

Large Language Models (LLMs) are increasingly applied to simulate human behavior across a range of domains~\citep{park2024generative, yang2024psychogat, suh2025language}, including education~\citep{macina2023mathdial,miroyan2025parastudent,ross2025modelingstudentlearning38} where student modeling has long been a foundational research problem in building tutoring systems. 
Recent approaches promise improved fidelity in predicting student actions~\citep{markel2023gpteach, khalil2025creating, xu2025classroom}, surpassing traditional approaches. 
However, LLM-based simulations are often limited to replicating surface-level patterns of user behavior, rather than modeling the latent processes underpinning observable outcomes~\citep{li2026simulating,wu2026humanlm}.

In educational settings, understanding the reasoning behind observed actions is crucial for applications such as diagnosing misconceptions, generating targeted feedback, and evaluating tutoring systems~\citep{brown1978diagnostic}. Two students may arrive at the same solution through entirely different reasoning processes, or produce the same incorrect answer for fundamentally different reasons. Research in psychology highlights the importance of accessing internal reasoning processes~\citep{stanovich2000advancing,johnson2010mental} and has proposed methods such as think-aloud protocols~\citep{ericsson1993protocol} to surface them: emphasizing that observable actions provide only a partial view of cognition, as similar behaviors can arise from very different underlying reasoning.

With LLMs, it is indeed feasible to elicit verbalization of internal reasoning prior to a generated action. 
Yet existing work on reasoning in LLMs has largely focused on improving correctness, encouraging models to produce logically consistent or factually accurate outputs~\citep{wei2022chain,yao2023tree,shao2024deepseekmath}. 
Human actions, however, are not always rational or correct: people frequently make errors, hold misconceptions, or apply incomplete strategies. 
In education especially, modeling such incorrect or partial reasoning is essential, as mistakes are a central part of the learning process.

\begin{figure*}[t]
    \centering
    \includegraphics[width=\linewidth]{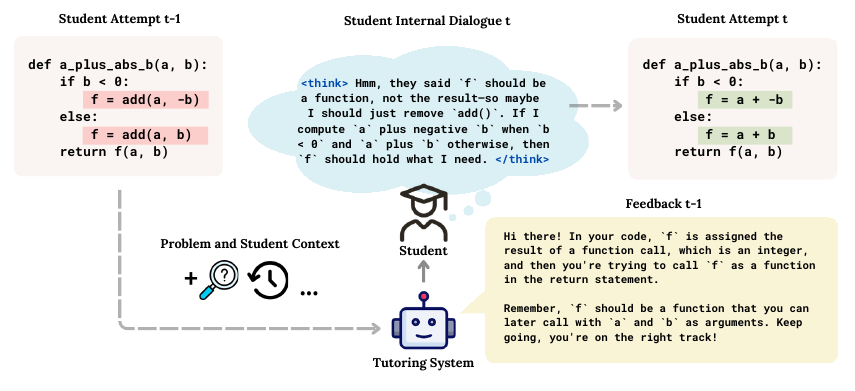}

    \caption{\textbf{\textsc{Internal Student Dialogue (INSIDE)}.} We propose INSIDE, a framework that fine-tunes LLMs to model both the \emph{internal reasoning} and the \emph{observable action} of students. In the context of AI tutor-assisted programming exercises, our trained LLM generates a verbalized chain-of-thought trajectory imitating how the student contextualizes previous code submissions and AI tutor feedback;  conditioned on the inferred internal dialogue, the model closely simulates the real student's code submission.}
    \label{fig:introduction}
\end{figure*}

To address this gap, we propose \textsc{\textbf{Internal Student Dialogue (INSIDE)}}, a student modeling framework that generates reasoning followed by action (\autoref{fig:introduction}). We argue that \textbf{the key affordance of student models lies in providing access to processes that are not directly observable in real students.} Unlike prior approaches that treat student behavior as a black box and focus solely on predicting outcomes, our framework explicitly models the latent reasoning process that precedes each action. Closest to our work, \citet{ross2025learning} models incorrect student reasoning by inferring misconceptions from erroneous answers—capturing the relationship between beliefs and errors, but doing so reconstructively (i.e., explaining post-hoc why an error occurred). In contrast, \textsc{INSIDE} introduces internal dialogue as an intermediate cognitive layer that precedes student actions during learning interactions, capturing the intent that leads to each attempt rather than only explaining mistakes after the fact. By modeling both \textit{how} and \textit{why} students act, \textsc{INSIDE} opens new directions for evaluation and human-centered optimization of tutoring systems.~\footnote{We release our code at \url{https://github.com/rosensh/inside}.}

In particular, our work makes the following contributions:

\begin{itemize}
\setlength\itemsep{2pt}

    \item \textbf{Reconstruction of pedagogically grounded reasoning traces.} We reconstruct latent reasoning traces from student interaction data via retrospective inference with a teacher model, conditioning on prior context and observed code edits. The resulting traces are grounded in educational theory and enable models to generate interpretable internal dialogue despite the absence of ground-truth reasoning.

    \item \textbf{A student modeling framework with internal dialogue.} We propose \textsc{INSIDE}, a framework that jointly models student code generation and the underlying reasoning process by conditioning actions on inferred internal dialogue.

    \item \textbf{A two-dimensional evaluation of simulation fidelity.} We evaluate our framework on (1) \emph{action fidelity}, the similarity between generated and real student code, and (2) \emph{reasoning quality}, defined as the alignment between generated reasoning and ground-truth code edits without requiring observed reasoning traces. We show that \textsc{INSIDE} improves action fidelity (lower Wasserstein distance) and achieves the highest reasoning alignment (up to 57.9\%) across models.

\end{itemize}

\section{Related Work}
\label{sec:related_work}

\subsection{LLM-Based Human and Student Simulation}
\label{subsec:related_work_llm_sim}

Recent literature proposes utilizing language models to simulate human behavior across various social sciences~\citep{simmons2022moral,hartmann2023political,wu2023large,bail2023we, Fatouros2024CanLL,horton2023large, serapio2023personality,hilliard2024eliciting,qian2025mask}.
By conditioning on virtual user personas, researchers have demonstrated LLMs can mimic patterns of human reasoning and social interaction across diverse contexts~\citep{park2023generative, park2024generative, santurkar2023whose, argyle2023out, suh2025language}.

Towards applications in education, researchers have similarly explored generative simulations of students \citep{markel2023gpteach, yue2024mathvc, khalil2025creating, xu2025classroom, tseng2024two}. 
Prior work ~\citep{miroyan2025parastudent} presents that fine-tuning LLMs on real student code trajectories improves models to capture realistic error patterns, stylistic variation, and incremental revision behavior.
However, this work still fails to emphasize modeling internal, latent reasoning of students that leads to replicating the observed actions, nor do the analyses measure fidelity and quality of the inferred internal dialogue by language models.
Our work directly addresses this gap by introducing structured internal reasoning traces grounded in pedagogical theory as a new training signal.

\subsection{LLM Reasoning for (In)Correct Responses}
Our proposed method models internal reasoning processes underlying learner behavior as a form of verbalized chain-of-thought (CoT)~\citep{wei2022chain}.
While the idea of CoT is widely researched as a means to improve the likelihood of producing correct, rational, or helpful responses from LLMs~\citep{yao2022react,kojima2022large,zhou2022least}, we note that this work concerns training models to generate CoT with fundamentally different goals and characteristics: replicating the potentially erroneous reasoning efforts naturally exhibiting uncertainty that real students would internally produce during the learning process.
In fact,  model post-training towards correct reasoning~\citep{shao2024deepseekmath,guo2025deepseek, yang2025qwen3} is in conflict with our goals of simulating human-like reasoning behavior, as reported in prior literature on human simulation~\citep{moon2024virtual, kang2025higher, wu2026humanlm}.
In \autoref{sec:dialogue}, we discuss the internal dialogue generation process, and in \autoref{sec:eval_internal_dialogue} we discuss evaluation criteria measuring quality and fidelity of the model-generated CoT in representing student-like internal thinking.

\subsection{Cognitive and Pedagogical Frameworks for Learning}
\label{subsec:related_work_pedagogy}
Bloom's Taxonomy \citep{bloom1956taxonomy, anderson2001taxonomy} provides a hierarchical classification of cognitive objectives—from basic recall through Apply, Analyze, Evaluate, and Create—and recent work finds that current benchmarks disproportionately probe lower-order skills \citep{huber2025llmstaxonomy}, motivating our explicit structuring of simulation data around these levels. Knowledge-tracing approaches, including Bayesian Knowledge Tracing \citep{corbett1994knowledge, anderson1995cognitive} and its deep learning successors \citep{piech2015deep} track \emph{what} students know but not \emph{why} they err; our fine-tuning approach trains LLMs to generate cognitively grounded reasoning traces with rich, explicit demonstration of the cognitive processes driving observable student behavior.

\section{Methodology}
\subsection{Data}
We study student code generation in an introductory programming course at the University of California, Berkeley, with $\sim$900 students per semester. The course spans core programming topics, including functions, recursion, sequences, trees, linked lists, and object-oriented programming. Each semester, students complete $\sim$10 homework assignments of 3--6 problems each. An autograder provides instant feedback on test case results, and students may resubmit freely until all tests pass, producing a multi-attempt submission stream for each student--problem pair. Beyond autograder feedback, each submission that fails at least one test case may also receive natural-language feedback from an LLM-based tutor~\citep{10.1145/3641554.3701864} if the student has consented to use the AI tutor. Every attempt is recorded, capturing the student's code, autograder results, and tutor feedback. Our data spans two semesters (Spring 2024 and Spring 2025) focusing on the first five homework assignments about Python programming. We partition the data by semester, using Spring 2025 for training (445 students, 2,022 submission streams, 6,911 total submissions) and Spring 2024 for testing (479 students, 1,546 submission streams, 6,316 total submissions). Further details regarding IRB and data privacy are provided in \autoref{sec:app-data}.

To evaluate generalization, we define two test subsets:

\textbf{\texttt{test\_OP} (Old Problems)} contains 5,262 code submissions from new test-semester students (Spring 2024) on problems that also appear in the training set. This set evaluates generalization to unseen students on familiar problems.

\textbf{\texttt{test\_NP} (New Problems)} contains 1,054 code submissions from Spring 2024 students solving problems not present in the training data. This set evaluates generalization to both unseen students and unseen problems.

When reporting problem-specific results, we select one representative problem from each subset with comparable difficulty, measured by the number of submissions per student submission stream: \texttt{test\_OP\_1} (968 submissions; 3.89 $\pm$ 4.48 submissions per stream) and \texttt{test\_NP\_1} (695 submissions; 3.46 $\pm$ 2.84 submissions per stream). Results for an additional representative problem from each subset, \texttt{test\_OP\_2} (759 submissions) and \texttt{test\_NP\_2} (359 submissions), are provided in \autoref{app:results}.

\subsection{Problem Formulation}
We formalize the problem of student code generation as follows: given a student $s_i$ (the $i$-th student) and a programming problem $p_u$ (the $u$-th problem), an LLM generates the next code submission at time step $t$, $c_{t,s_i, p_u}$ conditioned on the student’s prior submission history $\{c_{<t, s_i, p_u}\}$  and the corresponding natural-language feedback instances from an AI tutor $\{f_{<t, s_i, p_u}\}$. The model sees up to $k$ prior submissions ($k \leq 10$) with feedback context. Each student–problem pair is represented by a stream of sequential code submissions, from the first to the final attempt. 

\subsection{Internal Dialogue Generation}
\label{sec:dialogue}
We define the interaction context at time step $t$ as:
\[
x_t = \left(p_u,\; \{c_{<t, s_i, p_u},\; f_{<t, s_i, p_u}\}\right).
\]

At inference-time, a language model $\mathcal{M}$ generates internal dialogue $z_{t, s_i, p_u}$ that represents the reasoning process leading to the next code submission:
$c_{t, s_i, p_u}$:
\[
z_{t, s_i, p_u}
\sim
\mathcal{M}\big(
\cdot
\mid
x_t
\big),
\quad
c_{t, s_i, p_u}
\sim
\mathcal{M}\big(
\cdot
\mid
z_{t, s_i, p_u},\; x_t
\big).
\]

When generating the training data, we perform retrospective inference with a teacher model $\mathcal{T}$, where the teacher conditions on the interaction context $x_t$ and the observed (ground-truth, student-generated) next submission:
\[
z_{t, s_i, p_u}
\sim
\mathcal{T}\big(
\cdot
\mid
x_t,\;
c_{t, s_i, p_u}
\big).
\]

Generation proceeds in two stages. First, the teacher LLM infers the student's internal state in third person, producing structured summaries of each \textit{cognitive}, \textit{affective}, and \textit{action} states, inspired by three domains in Bloom’s Taxonomy~\citep{bloom1956taxonomy}. Next, given these inferred states, the model generates a first-person internal dialogue (a think trace) reflecting the student’s reasoning process prior to the observed submission. These generated internal dialogue traces approximate the student's latent reasoning process underlying the observed submission. We use GPT-5~\citep{gpt5} as the teacher model to generate these traces. The prompt used for generation is shown in \autoref{app:dialogue_prompt}.

\subsection{Experiments}
We evaluate the LLM's ability to generate student code under two settings.

\textbf{Experiment 1 (without CoT).}  
Generate the code submission at a given timestamp $t$ for student $s_i$ on problem $p_u$, conditioned on the student’s prior $k$ code submissions and their associated AI tutor feedback for the same problem, \emph{without intermediate CoT generation}. 

\textbf{Experiment 2 (with CoT).}  
Generate both the internal dialogue and the code submission at timestamp $t$ for student $s_i$ on problem $p_u$, conditioned on the student’s prior $k$ code submissions and their associated tutor feedback for the same problem. The model generates  $z_{t, s_i, p_u}$ and $c_{t, s_i, p_u}$ as a single sequence, with the reasoning trace produced before the code.

\subsection{Models}
We compare fine-tuning and prompting methods across both experiments. For fine-tuning, we train the model on the target outputs; in Experiment~2, this includes internal dialogue generated by the teacher model $\mathcal{T}$, which supervises the reasoning trace. For prompting, we provide the same interaction context and instruct the model to first generate a reasoning trace and then produce code. To isolate the effect of reasoning, we evaluate two prompting variants within Experiment~2: Experiment~2.1 uses standard CoT prompting, while Experiment~2.2 uses a structured CoT inspired by Bloom’s Taxonomy, aligning with the supervision used in fine-tuning. 

\begin{itemize}
\setlength\itemsep{2pt}
    \item \textbf{Fine-tuning.} We fine-tune Qwen2.5-7B~\citep{qwen2025qwen25technicalreport}, Qwen2.5-Coder-7B~\citep{qwen25_coder}, Qwen3-8B-Base~\citep{yang2025qwen3}, and LLaMA-3-8B~\citep{llama3} separately for Experiment~1 and 2 using LoRA~\citep{lora} ($r=16$, $\alpha=32$), for two epochs with a learning rate of $10^{-4}$. 
    \item \textbf{Prompting.} We evaluate instruction-tuned variants of each base model used for fine-tuning, including Qwen2.5-7B-Instruct, Qwen2.5-Coder-7B-Instruct, Qwen3-8B, and LLaMA-3-8B-Instruct, as well as GPT-5. Models are prompted with the same interaction context as fine-tuning inference and instructed to generate an internal reasoning trace followed by the next code submission. To study the effect of reasoning at scale, we additionally evaluate reasoning models Qwen3-14B and Qwen3-32B under prompting for Experiment~2.2 only.
\end{itemize}

Throughout the paper, we refer to models using experiment-specific naming conventions:
\begin{itemize}
\setlength\itemsep{2pt}
\item Fine-tuning
\begin{itemize}
\setlength\itemsep{2pt}
\item Experiment~1 models (w/o CoT): \{model\_name\}-SFT
\item Experiment~2 models (w/ CoT): \{model\_name\}-INSIDE
\end{itemize}
\item Prompting Baselines
\begin{itemize}
\setlength\itemsep{2pt}
\item Experiment~2.1 models (w/ CoT): \{model\_name\}-CoT
\item Experiment~2.2 models (w/ Bloom-Inspired CoT): \{model\_name\}-BloomCoT
\end{itemize}
\end{itemize}

Here, we mainly report results on Qwen models and GPT-5 in \autoref{sec:results}, and report additional results based on the LLaMA models in \autoref{app:results}. Details on the prompting templates are provided in \autoref{app:inference_prompt}. 

\section{Evaluation}

\subsection{Fidelity of Action}
\label{sec:eval_fidelity_of_action}
Because students' internal reasoning is not directly observable, we evaluate action fidelity through their code submissions, measuring how closely model-generated code matches real student submissions. For each student–problem instance, the model generates a single next-attempt submission conditioned on prior interaction history. We then compare the generated code against the real student code using the functionality and stylistic metrics introduced by \citet{miroyan2025parastudent}, which are drawn from metrics commonly used to assess student code in programming education. We report the Wasserstein distance (Earth Mover's Distance) for each metric to capture distributional similarity between generated and real student code, where lower values indicate closer alignment between distributions. 

\subsubsection{Code Functionality}
\label{sec:func}
We evaluate functional correctness using the same autograder test suite used for real student submissions. For each instance, the generated code is executed against the full test suite, and we report the fraction of tests passed (pass rate).

\subsubsection{Code Complexity and Style}
\label{sec:style}
To capture stylistic and complexity fidelity, we extract the following metrics:

\begin{itemize}
\setlength\itemsep{1pt}
    \item \textbf{Code length:} Measured by lines of code (LOC), reflecting verbosity.
    \item \textbf{Abstract Syntax Tree (AST) structure:} AST depth and width~\citep{ast}, capturing structural complexity.
    \item \textbf{PEP 8 violations:} Number of deviations from Python’s style guide (PEP 8)~\citep{pep8}, computed using \texttt{pycodestyle}\footnote{\url{https://pycodestyle.pycqa.org}}.
\end{itemize}
\vspace{-2mm}

\subsection{Quality of Internal Dialogue}
\label{sec:eval_internal_dialogue}
We evaluate the quality of the generated internal dialogue through \textbf{alignment}: whether the generated internal dialogue reflects the real student's code changes. Higher alignment suggests a closer approximation of the reasoning process underlying the student's observed behavior.
For each instance at time step $t$, we provide an LLM judge (GPT-5-mini) with the student's code $c_{t-1, s_i, p_u}$, the feedback $f_{t-1, s_i, p_u}$, the real student's next submission $c_{t, s_i, p_u}$, the model-generated internal dialogue $\hat{z}_{t, s_i, p_u}$, and the model-generated next submission $\hat{c}_{t, s_i, p_u}$. We compute code diffs between $c_{t-1, s_i, p_u}$ and $c_{t, s_i, p_u}$ to capture the ground-truth changes.
The judge decomposes the synthetic internal dialogue into a set of atomic claims $\mathcal{V}_{t} = \{v^{(1)}, \dots, v^{(n)}\}$ representing intended actions. Each claim is evaluated against the ground-truth diff using a binary indicator:
\[
\mathbb{1}_{\text{gt}}(v^{(i)}) =
\begin{cases}
1 & \text{if } v^{(i)} \text{ is reflected in } c_{t-1, s_i, p_u} \rightarrow c_{t, s_i, p_u} \\
0 & \text{otherwise}
\end{cases}
\]
We compute the fraction of supported claims:
\[
\texttt{Alignment}_t = \frac{1}{|\mathcal{V}_t|} \sum_{i=1}^{|\mathcal{V}_t|} \mathbb{1}_{\text{gt}}(v^{(i)}).
\]
We report alignment as proportions averaged across all instances. The prompt for the LLM-judge framework is provided in \autoref{app:eval-prompt}. To validate the labels assigned by the LLM judge, we evaluate it on the internal dialogue generated by the teacher model in \autoref{sec:dialogue}, paired with their corresponding code edits, where the judge should ideally assign full coverage. On a sample of code submissions from the training data ($n = 209$), the judge assigns an average alignment score of 95.2\% for transitions $c_{t-1, s_i, p_u} \rightarrow c_{t, s_i, p_u}$ given $z_{t, s_i, p_u}$, indicating that the metric reliably attributes stated reasoning to observed code changes. Furthermore, we randomly sampled 25 generated internal dialogues across all models and manually annotated whether each extracted claim was reflected in the corresponding code edit. The manual annotations achieved 88.0\% agreement with the LLM judge labels ($\kappa = 0.754$), indicating substantial agreement between the LLM judge and human annotations.

\section{Results}
\label{sec:results}

\subsection{Action Fidelity}
\label{sec:fidelity}
\begin{table*}[h]
\centering
\caption{\textbf{Summary of distributional similarity between model-generated and student code.} Each cell reports the Wasserstein distance between the distribution of model-generated outputs and real student submissions for each metric. 
Metrics span pass rate, LOC, AST depth and width, and PEP 8 violations. Models fine-tuned with INSIDE produce code distributions closest to student data on \texttt{test\_OP}, and maintain comparable alignment to SFT on \texttt{test\_NP}.  
Bolded and underlined values indicate the lowest (best) Wasserstein distances.}
\label{tab:wasserstein}

\begin{subtable}[h]{0.48\textwidth}
\centering
\caption{\texttt{test\_OP}} 
\label{tab:wasserstein-a}
\adjustbox{max width=\textwidth}{%
\renewcommand{\arraystretch}{1.25}
\begin{tabular}{lccccc}
\toprule
\multirow{2}{*}{\textbf{Model}} 
& \textbf{Pass Rate} $\downarrow$
& \textbf{LOC} $\downarrow$ 
& \multicolumn{2}{c}{\textbf{AST} $\downarrow$} 
& \textbf{PEP 8} $\downarrow$ \\
\cmidrule(lr){4-5}
& & & \textbf{Depth} & \textbf{Width} & \\
\midrule
GPT-5 & 0.60 & 0.74 & 0.54 & 1.08 & 1.29 \\
Qwen2.5-7B-Instruct & 0.39 & 0.71 & 0.60 & 0.98 & 1.12 \\
Qwen2.5-Coder-7B-Instruct & 0.46 & 0.80 & 0.68 & 1.78 & 1.13 \\
Qwen3-8B & 0.44 & 1.05 & 0.54 & 1.33 & 1.38 \\
\hb Qwen2.5-7B-SFT & 0.14 & 0.29 & 0.29 & 0.52 & 0.18 \\
\hb Qwen2.5-Coder-7B-SFT & 0.15 & 0.29 & 0.29 & 0.51 & 0.19 \\
\hb Qwen3-8B-SFT & 0.15 & 0.28 & 0.28 & 0.55 & \underline{\textbf{0.16}} \\
\midrule
GPT-5-CoT & 0.61 & 1.10 & 0.72 & 1.20 & 1.37 \\
Qwen2.5-7B-Instruct-CoT & 0.38 & 0.67 & 0.60 & 0.98 & 1.16 \\
Qwen2.5-Coder-7B-Instruct-CoT & 0.47 & 0.93 & 0.75 & 1.50 & 1.25 \\
Qwen3-8B-CoT & 0.48 & 1.08 & 0.59 & 1.46 & 1.38 \\
\midrule
GPT-5-BloomCoT & 0.60 & 1.24 & 0.69 & 1.19 & 1.38 \\
Qwen2.5-7B-Instruct-BloomCoT & 0.35 & 0.67 & 0.58 & 0.90 & 1.17 \\
Qwen2.5-Coder-7B-Instruct-BloomCoT & 0.43 & 0.88 & 0.71 & 1.40 & 1.19 \\
Qwen3-8B-BloomCoT & 0.47 & 0.98 & 0.54 & 1.30 & 1.37 \\
\hy Qwen2.5-7B-INSIDE & \underline{\textbf{0.05}} & 0.26 & 0.27 & \underline{\textbf{0.39}} & 0.18 \\
\hy Qwen2.5-Coder-7B-INSIDE & \underline{\textbf{0.05}} & \underline{\textbf{0.21}} & 0.21 & 0.40 & \underline{\textbf{0.16}} \\
\hy Qwen3-8B-INSIDE & \underline{\textbf{0.05}} & 0.28 & \underline{\textbf{0.20}} & 0.42 & 0.21 \\
\bottomrule
\end{tabular}}
\end{subtable}\hfill
\begin{subtable}[h]{0.48\textwidth}
\centering
\caption{\texttt{test\_NP}}
\label{tab:wasserstein-b}
\adjustbox{max width=\textwidth}{%
\renewcommand{\arraystretch}{1.25}
\begin{tabular}{lccccc}
\toprule
\multirow{2}{*}{\textbf{Model}} 
& \textbf{Pass Rate} $\downarrow$
& \textbf{LOC} $\downarrow$ 
& \multicolumn{2}{c}{\textbf{AST} $\downarrow$} 
& \textbf{PEP 8} $\downarrow$ \\
\cmidrule(lr){4-5}
& & & \textbf{Depth} & \textbf{Width} & \\
\midrule
GPT-5 & 0.52 & 0.64 & 1.24 & 2.72 & 1.36 \\
Qwen2.5-7B-Instruct & 0.37 & 0.86 & 0.50 & 1.05 & 0.82 \\
Qwen2.5-Coder-7B-Instruct & 0.45 & 0.85 & 0.52 & 0.71 & 1.01 \\
Qwen3-8B & 0.45 & 1.10 & 0.61 & 1.05 & 1.13 \\
\hb Qwen2.5-7B-SFT & \underline{\textbf{0.01}} & 0.20 & 0.19 & 0.28 & 0.14 \\
\hb Qwen2.5-Coder-7B-SFT & 0.03 & 0.35 & 0.18 & 0.39 & 0.16 \\
\hb Qwen3-8B-SFT & 0.02 & 0.23 & \underline{\textbf{0.17}} & 0.28 & \underline{\textbf{0.12}} \\
\midrule
GPT-5-CoT & 0.53 & 0.78 & 1.33 & 2.56 & 1.41 \\
Qwen2.5-7B-Instruct-CoT & 0.36 & 0.90 & 0.52 & 1.21 & 1.00 \\
Qwen2.5-Coder-7B-Instruct-CoT & 0.43 & 0.81 & 0.57 & 0.85 & 1.06 \\
Qwen3-8B-CoT & 0.44 & 1.06 & 0.74 & 1.37 & 1.21 \\
\midrule
GPT-5-BloomCoT & 0.51 & 0.94 & 1.12 & 1.86 & 1.42 \\
Qwen2.5-7B-Instruct-BloomCoT & 0.33 & 0.84 & 0.47 & 0.96 & 0.95 \\
Qwen2.5-Coder-7B-Instruct-BloomCoT & 0.41 & 0.82 & 0.58 & 1.13 & 0.92 \\
Qwen3-8B-BloomCoT & 0.42 & 1.01 & 0.68 & 1.19 & 1.18 \\
\hy Qwen2.5-7B-INSIDE & 0.04 & \underline{\textbf{0.16}} & 0.22 & 0.37 & 0.18 \\
\hy Qwen2.5-Coder-7B-INSIDE & 0.05 & 0.20 & 0.18 & \underline{\textbf{0.14}} & 0.19 \\
\hy Qwen3-8B-INSIDE & 0.04 & 0.20 & 0.19 & 0.40 & 0.18 \\
\bottomrule
\end{tabular}}
\end{subtable}

\end{table*}
\begin{figure*}[h]
    \centering

    \begin{subfigure}[h]{1.0\textwidth}
        \centering
        \includegraphics[width=\linewidth]{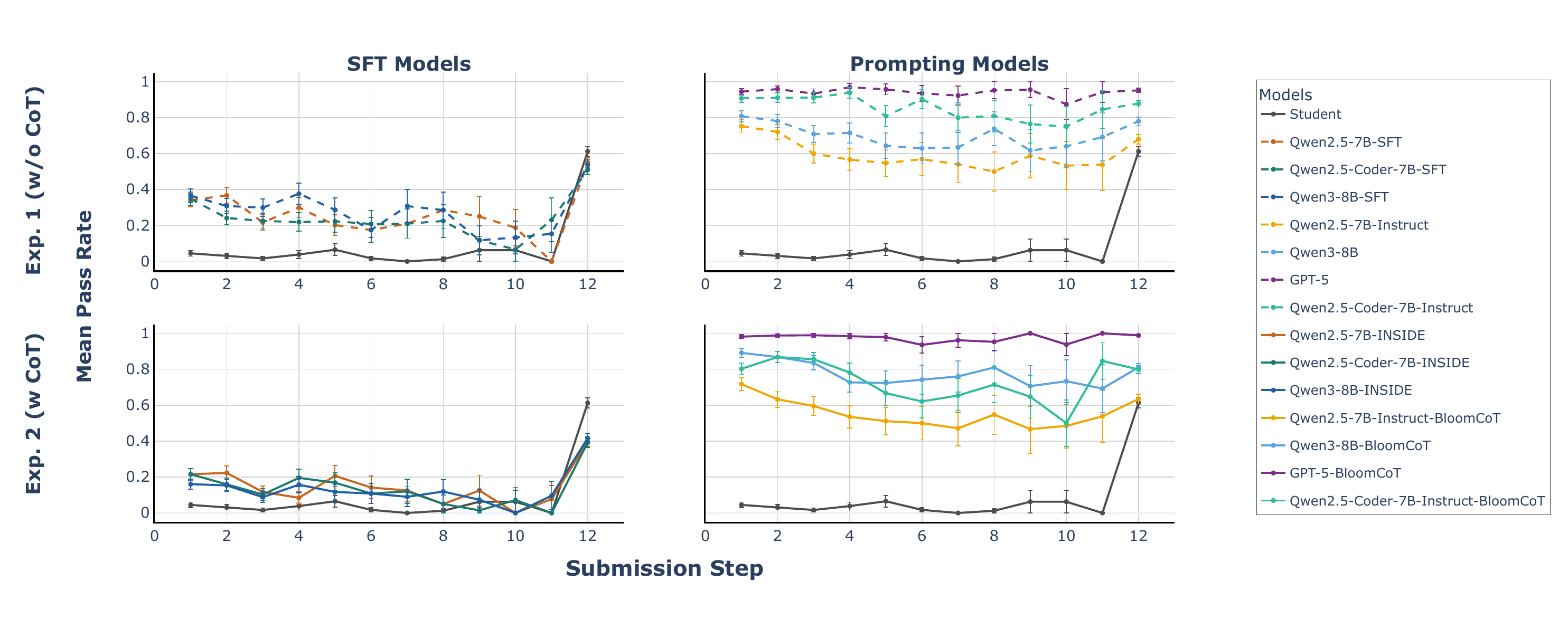}
        \caption{\texttt{test\_OP\_1} (Representative Problem: $n=968$)}
        \label{fig:pass-a}
    \end{subfigure}
    
    \vspace{0.8em}
    
    \begin{subfigure}[h]{1.0\textwidth}
        \centering 
        \includegraphics[width=\linewidth]{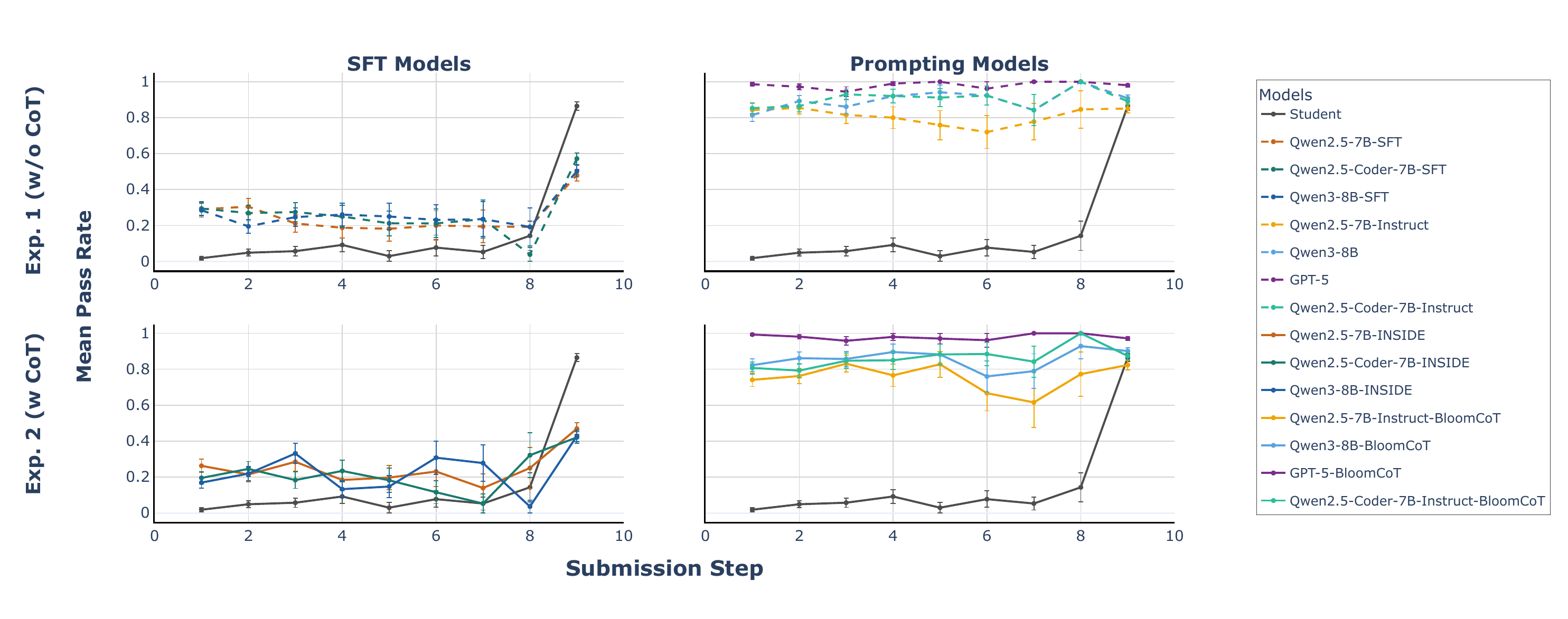}
        \caption{\texttt{test\_NP\_1} (Representative Problem: $n=695$)}
        \label{fig:pass-b}
    \end{subfigure}
    
\caption{\textbf{Pass rate trajectories over submission steps.} Submission steps correspond to attempt indices;  steps are aligned such that final submissions map to a shared terminal step, enabling comparison across trajectories of different lengths. Real students exhibit low initial pass rates, followed by a sharp increase near the final step, reflecting incremental problem-solving. Fine-tuned models capture this pattern, and we observe \textsc{INSIDE} further improves alignment with real student patterns. In contrast, prompting-based approaches consistently maintain artificially high pass rates.}

\label{fig:pass}
\end{figure*}

To measure distributional similarity between model-generated and student code, we use the Wasserstein distance over each evaluation metric. For each metric, we compare the distribution of generated outputs against the real student distribution. We estimate Wasserstein distance via bootstrap resampling (500 resamples) and report the results in \autoref{tab:wasserstein}.

Across both settings, fine-tuned models substantially outperform prompting-based methods, and within prompting approaches, more structured strategies improve alignment: BloomCoT generally achieves lower Wasserstein distances, though it still underperforms compared to fine-tuning.

On \texttt{test\_OP} (\autoref{tab:wasserstein-a}), \textsc{INSIDE} consistently achieves the lowest Wasserstein distances across all metrics, indicating the closest match to real student code distributions. Compared to the SFT baseline, incorporating internal dialogue improves alignment not only in functionality but also in stylistic and structural properties of code. On \texttt{test\_NP} (\autoref{tab:wasserstein-b}), results are more mixed: models fine-tuned with regular SFT and \textsc{INSIDE} perform comparably across most metrics, with smaller Wasserstein distances compared to \texttt{test\_OP}. 

One possible explanation for the smaller improvement of \textsc{INSIDE} over SFT on \texttt{test\_NP} is that the two evaluation splits have different student pass-rate distributions. In particular, \texttt{test\_NP} contains a higher proportion of successful student submissions than \texttt{test\_OP}. This matters because fine-tuning on student trajectories already substantially reduces the over-competence bias observed in prompted models, bringing SFT closer to the target student distribution. When student failures are more common, as in \texttt{test\_OP}, this remaining mismatch is larger, so \textsc{INSIDE} has more room to improve by better matching the student pass-rate distribution. When student successes are more common, as in \texttt{test\_NP}, SFT is already close to the target distribution, leaving less mismatch for \textsc{INSIDE} to correct. We provide a detailed distributional analysis in \autoref{app:pass_dist}.


\subsubsection{Action Fidelity Results on Representative Problem}

\autoref{fig:pass} shows how pass rates evolve for model-generated code compared to real students. In this analysis, we focus on the effect of reasoning by comparing models with and without internal dialogue (CoT vs.\ no CoT). For prompting-based methods, we report results using BloomCoT, as it achieves stronger action fidelity overall than standard CoT, as shown in Section~5.1. Pass rate serves as a primary indicator of student progress in our setting, where students iteratively interact with AI feedback and make incremental code revisions toward a correct solution. Because these dynamics are problem-dependent, we report representative results on one problem from each setting: \texttt{test\_OP\_1} and \texttt{test\_NP\_1}. Across both problems, real students start with low pass rates and exhibit a sharp increase near the final steps, reflecting incremental progress. SFT models closely track this trajectory. In contrast, prompting-based models maintain relatively high pass rates ($\approx 80\%$) from the beginning, with little variation across steps.

To quantify alignment with student behavior, we compute the Mean Absolute Error (MAE) between model and student pass rates, averaged over the steps within each experiment. On \texttt{test\_OP\_1}, the top three models are Qwen3-8B-INSIDE (0.094), Qwen2.5-Coder-7B-INSIDE (0.098), and Qwen2.5-7B-INSIDE (0.113). On \texttt{test\_NP\_1}, the closest models are Qwen2.5-Coder-7B-INSIDE (0.162), Qwen2.5-7B-SFT (0.181), and Qwen2.5-7B-INSIDE (0.182). This suggests that conditioning on inferred internal dialogue may enable the model to better reason and reflect on received AI tutor feedback before generating code, contributing to closer alignment with the incremental progress observed in student pass rate trajectories.
\begin{table}[h]
\centering
\caption{\textbf{Alignment between generated internal dialogue and ground-truth code edits.} 
Each cell reports alignment (\%) with standard error in parentheses. 
Results are shown for models that generate internal dialogue (Experiment~2). 
INSIDE achieves the highest alignment across both settings, indicating closer correspondence between generated reasoning and observed code edits. Higher values indicate better alignment. Bolded and underlined values denote the best-performing model in each setting.}
\label{tab:think}
\begin{subtable}[h]{0.48\textwidth}
\centering
\caption{\texttt{test\_OP}}
\label{tab:align_op}
\adjustbox{max width=\linewidth}{%
\renewcommand{\arraystretch}{1.2}
\begin{tabular}{l c}
\toprule
\textbf{Model} & \textbf{\texttt{Alignment} (\%)} \\
\midrule
GPT-5-BloomCoT                        & 45.5 (0.8) \\
Qwen2.5-7B-Instruct-BloomCoT          & 49.6 (0.9) \\
Qwen2.5-Coder-7B-Instruct-BloomCoT    & 46.0 (0.8) \\
Qwen3-8B-BloomCoT            & 47.9 (0.8) \\
Qwen3-14B-BloomCoT                    & 46.5 (0.8) \\
Qwen3-32B-BloomCoT                    & 44.4 (0.8) \\
\midrule
\hy Qwen2.5-7B-INSIDE                     & \underline{\textbf{51.8}} (0.8) \\
\hy Qwen2.5-Coder-7B-INSIDE               & 49.2 (0.8) \\
\hy Qwen3-8B-INSIDE                       & 50.5 (0.8) \\
\bottomrule
\end{tabular}%
}
\end{subtable}
\hfill
\begin{subtable}[h]{0.48\textwidth}
\centering
\caption{\texttt{test\_NP}}
\label{tab:align_np}
\adjustbox{max width=\linewidth}{%
\renewcommand{\arraystretch}{1.2}
\begin{tabular}{l c}
\toprule
\textbf{Model} & \textbf{\texttt{Alignment} (\%)} \\
\midrule
GPT-5-BloomCoT                        & 53.5 (2.1) \\
Qwen2.5-7B-Instruct-BloomCoT          & 55.1 (2.2) \\
Qwen2.5-Coder-7B-Instruct-BloomCoT    & 51.3 (2.2) \\
Qwen3-8B-BloomCoT            & 56.0 (2.3) \\
Qwen3-14B-BloomCoT                    & 55.6 (2.3) \\
Qwen3-32B-BloomCoT                    & 52.4 (2.2) \\
\midrule
\hy Qwen2.5-7B-INSIDE                     & 56.9 (2.0) \\
\hy Qwen2.5-Coder-7B-INSIDE               & 55.5 (2.0) \\
\hy Qwen3-8B-INSIDE                       & \underline{\textbf{57.9}} (2.0) \\
\bottomrule
\end{tabular}%
}
\end{subtable}
\end{table}

\subsection{Quality of Internal Dialogue}
\label{sec:quality_of_internal_dialogue}  
\autoref{tab:think} shows alignment score (\autoref{sec:eval_internal_dialogue}) between generated internal dialogue and ground-truth code edits. This evaluation is restricted to models that generate CoT before code generation (Experiment~2). 
Overall, \textsc{INSIDE} achieves the highest alignment across both settings. On \texttt{test\_OP}, Qwen2.5-7B-INSIDE reaches 51.8\%, outperforming the best prompting baseline, Qwen2.5-7B-Instruct-BloomCoT. On \texttt{test\_NP}, Qwen3-8B-INSIDE achieves 57.9\% compared to 56.0\% for the strongest BloomCoT model. Although scores on \texttt{test\_NP} are higher overall, this split contains substantially fewer samples, resulting in larger standard errors ($\approx$ 2.0–2.3 vs. $\approx$ 0.8–0.9), which limits direct comparison across settings. Interestingly, larger and more capable models (e.g., GPT-5 and Qwen3-32B) tend to achieve lower alignment scores, suggesting that stronger reasoning ability does not necessarily translate to reasoning that matches student-like code edits. We additionally report self-consistency results, measuring alignment between generated internal dialogue and model-generated code, in \autoref{app:self-consistency}.

Furthermore, alignment should be interpreted jointly with action fidelity. High alignment alone does not indicate realistic student modeling if the generated code does not follow plausible solution trajectories. This is a challenging task: the model must generate reasoning that aligns with specific code edits, rather than producing plausible but generic explanations. While some prompting-based models show alignment comparable to \textsc{INSIDE} (e.g., Qwen2.5-7B-Instruct-BloomCoT on \texttt{test\_OP}), as shown in \autoref{sec:fidelity}, these models exhibit poor action fidelity and unrealistic solution trajectories, indicating a disconnect between explanation and behavior. In contrast, \textsc{INSIDE} achieves both high alignment and strong action fidelity, producing reasoning that is consistent with the observed edits while also generating code that follows realistic student progression. This joint improvement suggests that internal dialogue helps bridge the gap between explanation and behavior, rather than optimizing for alignment alone. We provide qualitative examples of model-generated internal dialogue in Appendix~\ref{app:examples}.

\section{Limitations and Future Work}
A key limitation of our approach is that internal dialogue is reconstructed rather than observed. The reasoning traces are generated by a teacher LLM through retrospective inference, and therefore represent an approximation of the student’s latent cognition. Importantly, because LLMs are typically trained to produce expert-like reasoning, they may struggle to faithfully reconstruct novice reasoning patterns, even in a reconstruction setting. As a result, the generated traces may reflect more coherent or structured reasoning than what real students exhibit. While this provides a useful proxy, it remains a best-effort estimate of what the student might have been thinking. Future work can further validate and calibrate these reconstructed traces through human-centered methods such as think-aloud studies or retrospective verbalization protocols. 

Our evaluation also has a distributional limitation: the splits between \texttt{test\_OP} and \texttt{test\_NP} differ not only in whether problems are seen or unseen, but also in their underlying student pass-rate distributions, which can affect the interpretation of the results. Additionally, while \textsc{INSIDE} achieves the highest alignment among models (reaching $\sim$58\%), this shows that a majority of generated claims explain student code edits, with remaining gaps indicating room for further improvement. 

Moreover, compared to the near-perfect alignment of teacher-generated traces, this suggests opportunities to better capture reasoning that consistently accounts for student actions. Future work can explore alternative approaches to reasoning generation, such as reinforcement learning methods~\citep{wu2026humanlm}, which encourage reasoning to emerge through reward modeling rather than supervised fine-tuning.

Despite these limitations, incorporating internal dialogue does not degrade action fidelity compared to SFT. On \texttt{test\_OP}, \textsc{INSIDE} more closely matches student behavior, and on \texttt{test\_NP}, performance remains comparable. This highlights a key affordance of student simulators: beyond replicating behavior, they provide access to otherwise unobservable signals—internal dialogue. This enables new applications in educational systems. Student simulators are increasingly used to evaluate and optimize AI tutors prior to deployment~\citep{dinucu2025problem}. By modeling both actions and reasoning, \textsc{INSIDE} allows evaluating whether feedback resolves misconceptions and supports counterfactual analysis of alternative interventions. Beyond evaluation, \textsc{INSIDE} enables learner-facing tools. Externalizing internal dialogue can support reflection and metacognition~\citep{kumar2024supporting}, and provide richer student representations for clustering based on reasoning patterns rather than noisy surface behavior~\citep{lu2025improve}.

\section{Conclusion}
We presented \textsc{INSIDE}, a student simulation framework that jointly models student actions and internal reasoning. Incorporating internal dialogue improves alignment with student behavior and achieves the highest reasoning quality among methods. Thus, modeling reasoning beyond observable actions is key to building more realistic student simulators and enables new opportunities for evaluating and improving tutoring systems.
Beyond observed outcome simulation fidelity, \textsc{INSIDE} points to LLM-based simulations grounded in cognitively plausible internal reasoning, presenting opportunities for developing tutoring systems equipped with misconception-aware interventions. 


\bibliographystyle{colm2026_conference}
\bibliography{inside}

\appendix
\appendix
\clearpage

\section{Data Privacy}
\label{sec:app-data}
We received IRB approval for the research use of the dataset (protocol ID: 2023-09-16725), after which the AI tutor was deployed in the course. When students submitted their code to the autograder as part of the regular course practice, they were asked whether they would like to receive AI tutor feedback on their code, and if so, were asked whether they would like their data to be used for research. Students were still able to receive AI feedback even if they did not wish their data to be used in research. Only students who explicitly consented were included in our analysis. Additionally, all student identifiers (e.g., IDs and emails) were fully anonymized prior to passing the data to the LLM and any further analysis. As part of our system design, any data from LLM calls from our educational platform to the model providers will not be retained on third-party servers and will not be used for further fine-tuning of those models. 

\section{Extended Action Fidelity Results}
\begin{figure*}[h]
    \centering

    \begin{subfigure}[h]{1.0\textwidth}
        \centering
        \includegraphics[width=\linewidth]{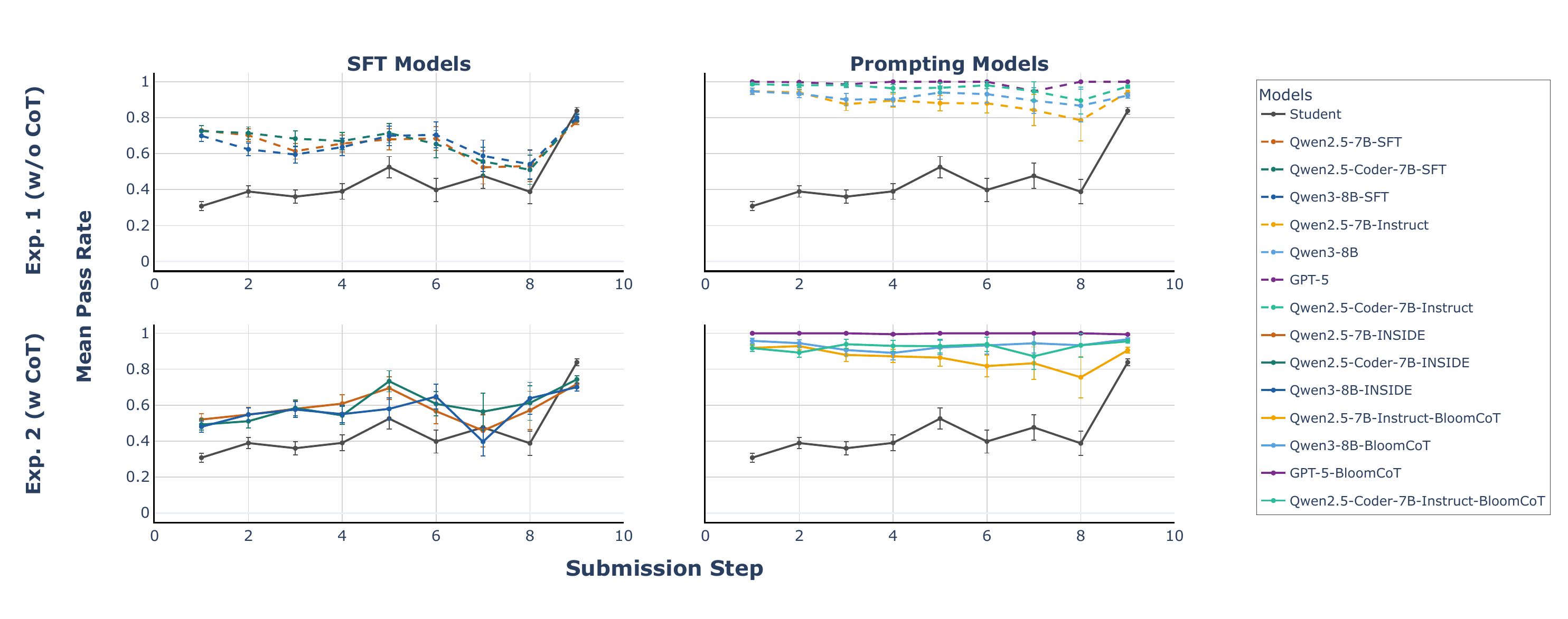}
        \caption{\texttt{test\_OP\_2} (Representative Problem: $n=759$)}
        \label{fig:pass-a}
    \end{subfigure}
    
    \vspace{0.8em}
    
    \begin{subfigure}[h]{1.0\textwidth}
        \centering 
        \includegraphics[width=\linewidth]{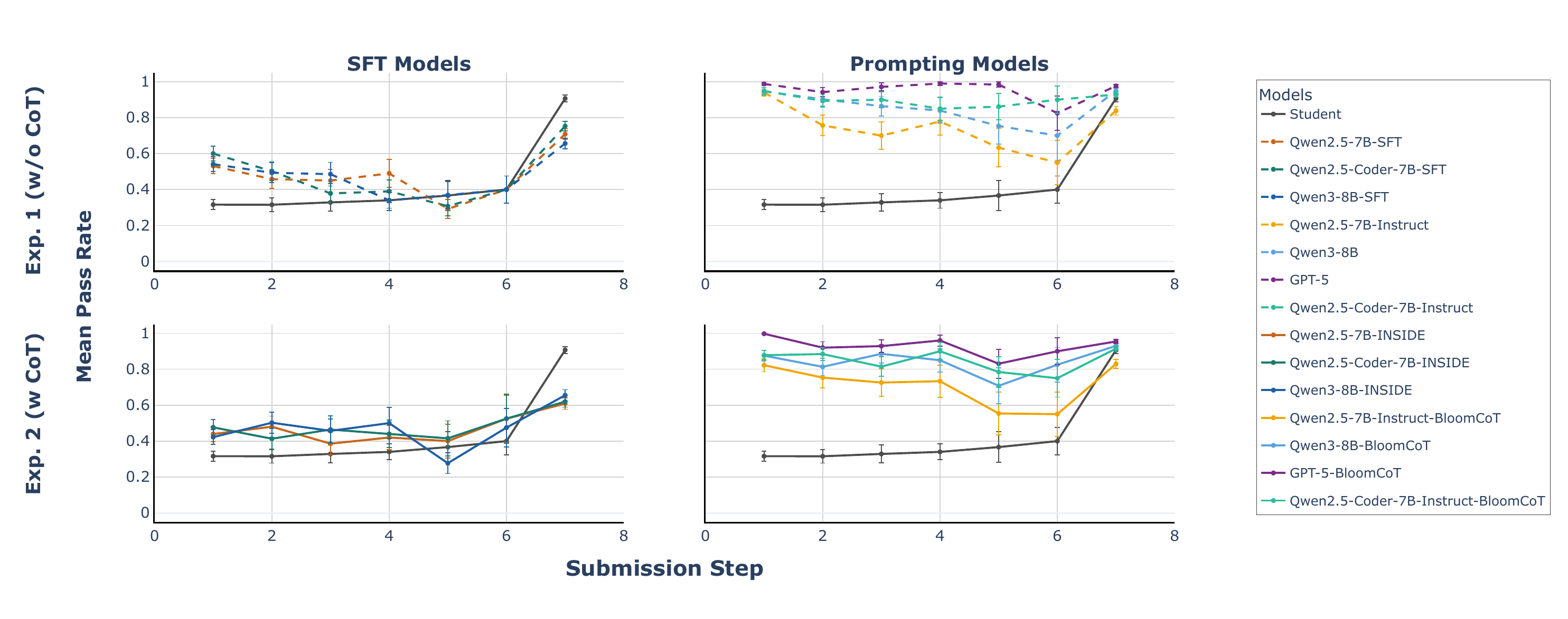}
        \caption{\texttt{test\_NP\_2} (Representative Problem: $n=359$)}
        \label{fig:pass-b}
    \end{subfigure}
    
\caption{\textbf{Pass rate trajectories over normalized submission steps on \texttt{test\_OP\_2} and \texttt{test\_NP\_2}.}}

\label{fig:app-pass}
\end{figure*}

This section provides extended results that complement the main paper. First, we report results on \texttt{test\_OP\_2} and \texttt{test\_NP\_2} to evaluate model performance on additional test problems. The progression of pass rate is shown in \autoref{fig:app-pass}, exhibiting trends consistent with \texttt{test\_OP\_1} and \texttt{test\_NP\_1}. Second, we report additional results for LLaMA and its Wasserstein distance metrics in \autoref{app:llama}.

\label{app:results}
\begin{table}[h]
\centering
\caption{\textbf{LLaMA results (Wasserstein distances)}.}
\label{app:llama}
\small
\setlength{\tabcolsep}{4pt}
\begin{tabular}{lccccc}
\toprule
\textbf{Model} & \textbf{Pass Rate}~$\downarrow$ & \textbf{LOC}~$\downarrow$ & \textbf{AST Depth}~$\downarrow$ & \textbf{AST Width}~$\downarrow$ & \textbf{PEP 8}~$\downarrow$ \\
\midrule
LLaMA-3-8B-Instruct & 0.20 & 0.99 & 0.52 & 1.12 & 0.98 \\
LLaMA-3-8B-Instruct-CoT & 0.18 & 0.85 & 0.48 & 1.00 & 1.01 \\
LLaMA-3-8B-Instruct-BloomCoT & 0.17 & 0.78 & 0.45 & 1.03 & 1.16 \\
LLaMA-3-8B-SFT & 0.07 & 0.32 & \underline{\textbf{0.17}} & \underline{\textbf{0.37}} & \underline{\textbf{0.20}} \\
LLaMA-3-8B-INSIDE & \underline{\textbf{0.06}} & \underline{\textbf{0.26}} & 0.22 & 0.48 & 0.21 \\
\bottomrule
\end{tabular}
\end{table}

\section{Self-Consistency of the Models}
\label{app:self-consistency}

\autoref{tab:app_self} reports self-consistency, a variant of the alignment metric described in \autoref{sec:eval_internal_dialogue}, measuring alignment between generated internal dialogue and the model's own generated code. Formally, we replace the ground-truth transition $c_{t-1,s_i,p_u} \rightarrow c_{t,s_i,p_u}$ with the model-generated transition $c_{t-1,s_i,p_u} \rightarrow \hat{c}_{t,s_i,p_u}$ and compute
\[
\texttt{SelfConsistency}_t =
\frac{1}{|\mathcal{V}_t|}
\sum_{i=1}^{|\mathcal{V}_t|}
\mathbb{1}_{\mathrm{self}}(v^{(i)}),
\]
where $\mathbb{1}_{\mathrm{self}}(v^{(i)})=1$ if claim $v^{(i)}$ is reflected in the model-generated code diff and $0$ otherwise.

Prompting-based models achieve high self-consistency scores (86.9\%--99.0\% across both test sets), with GPT-5 attaining the highest scores. However, as shown in our action fidelity evaluation (\autoref{sec:fidelity}), GPT-5 performs among the worst in terms of reproducing realistic student code trajectories. Taken together, these results suggest that strong consistency between a model's reasoning and its own generated code does not necessarily imply alignment with real student reasoning or behavior. While prompting-based methods naturally exhibit higher self-consistency due to explicit CoT prompting, \textsc{INSIDE} also achieves competitive self-consistency (83.0\%--87.3\% across both test sets), indicating that its generated reasoning remains coherent with its generated code while better matching real student code edits.

\begin{table}[h]
\centering
\caption{\textbf{Self-consistency between generated internal dialogue and model-generated code.} 
Each cell reports self-consistency (\%) with standard error in parentheses.
Higher values indicate better agreement between the generated reasoning and the model's own generated code.}
\label{tab:app_self}
\begin{subtable}[h]{0.48\textwidth}
\centering
\caption{\texttt{test\_OP}}
\label{tab:self_op}
\adjustbox{max width=\linewidth}{%
\renewcommand{\arraystretch}{1.2}
\begin{tabular}{l c}
\toprule
\textbf{Model} & \textbf{Self-consistency (\%)} \\
\midrule
GPT-5-BloomCoT                        & 97.6 (0.3) \\
Qwen2.5-7B-Instruct-BloomCoT          & 86.9 (0.6) \\
Qwen2.5-Coder-7B-Instruct-BloomCoT    & 89.6 (0.5) \\
Qwen3-8B-BloomCoT            & 91.9 (0.5) \\
Qwen3-14B-BloomCoT                    & 92.3 (0.5) \\
Qwen3-32B-BloomCoT                    & 90.9 (0.5) \\
\midrule
Qwen2.5-7B-INSIDE                     & 85.0 (0.6) \\
Qwen2.5-Coder-7B-INSIDE               & 86.3 (0.6) \\
Qwen3-8B-INSIDE                       & 83.0 (0.7) \\
\bottomrule
\end{tabular}%
}
\end{subtable}
\hfill
\begin{subtable}[h]{0.48\textwidth}
\centering
\caption{\texttt{test\_NP}}
\label{tab:self_np}
\adjustbox{max width=\linewidth}{%
\renewcommand{\arraystretch}{1.2}
\begin{tabular}{l c}
\toprule
\textbf{Model} & \textbf{Self-consistency (\%)} \\
\midrule
GPT-5-BloomCoT                        & 99.0 (0.3) \\
Qwen2.5-7B-Instruct-BloomCoT          & 90.3 (1.4) \\
Qwen2.5-Coder-7B-Instruct-BloomCoT    & 91.7 (1.2) \\
Qwen3-8B-BloomCoT            & 93.2 (1.1) \\
Qwen3-14B-BloomCoT                    & 91.3 (1.3) \\
Qwen3-32B-BloomCoT                    & 92.2 (1.3) \\
\midrule
Qwen2.5-7B-INSIDE                     & 85.3 (1.6) \\
Qwen2.5-Coder-7B-INSIDE               & 87.3 (1.4) \\
Qwen3-8B-INSIDE                       & 83.8 (1.6) \\
\bottomrule
\end{tabular}%
}
\end{subtable}
\end{table}

\clearpage
\section{Student Pass-Rate Distribution Across Test Splits}
\label{app:pass_dist}

\begin{figure}[h]
    \centering
    \includegraphics[width=0.85\textwidth]{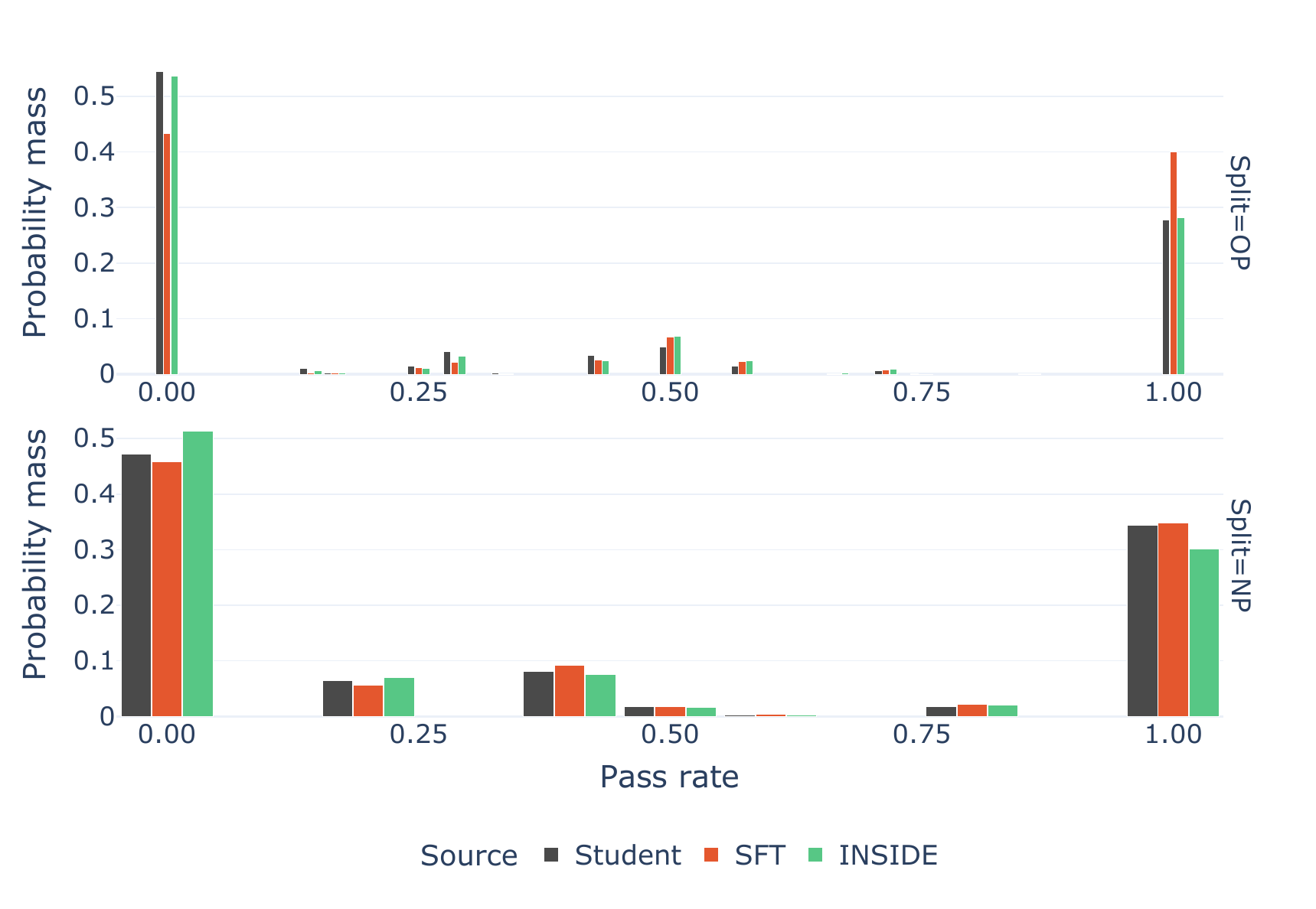}
    \caption{\textbf{Exact pass-rate mass by split: Student vs.\ SFT vs.\ INSIDE.} Each
    panel shows the probability mass of student, SFT, and INSIDE submissions at each
    exact pass rate, faceted by test split (\texttt{test\_OP} top, \texttt{test\_NP}
    bottom). Both splits show student distributions concentrated at the endpoints
    (pass rate 0 or 1), but differ in their failure/success ratio, which helps explain
    the asymmetric INSIDE--SFT gap across splits.}
    \label{fig:pass_dist}
\end{figure}

We compare empirical student and model pass-rate distributions in \autoref{fig:pass_dist}, since pass rate directly captures whether models generate submissions at student-like correctness levels.

Both \texttt{test\_OP} and \texttt{test\_NP} student distributions are concentrated at the endpoints, with 83\% and 81\% of mass at pass rates 0 and 1, respectively. However, their failure/success balance differs: \texttt{test\_OP} students fail at 55\% and succeed at 28\% (roughly 2:1), whereas \texttt{test\_NP} students fail at 47\% and succeed at 34\% (roughly 1.4:1). This helps explain the smaller \textsc{INSIDE}--SFT gap on \texttt{test\_NP}. On \texttt{test\_OP}, SFT underestimates complete failure (43\% vs. 55\%) and overestimates complete success (40\% vs. 28\%), reflecting residual over-competence from the base model; \textsc{INSIDE} better corrects this mismatch. On \texttt{test\_NP}, where students fail less and succeed more, SFT is already closer to the student distribution, leaving less mismatch for \textsc{INSIDE} to correct.

This also helps explain why Wasserstein distances are smaller on \texttt{test\_NP}, especially for SFT. \autoref{fig:pass_dist} suggests that SFT's smaller pass-rate Wasserstein distances on \texttt{test\_NP} are driven largely by its close match to fully passing submissions. Since fully correct solutions in introductory Python problems often share similar structure, including LOC, AST shape, and style, SFT's calibration at pass rate 1.0 can also reduce distances on related action-fidelity metrics.

\newpage
\section{Internal Dialogue Generation Prompt}

\label{app:dialogue_prompt}

We provide the full prompt used to generate synthetic internal dialogue with the teacher model in \autoref{fig:dialogue-prompt}. The prompt performs retrospective inference over real student prior code, feedback, and the observed next submission, producing structured cognitive, affective, and action states followed by a first-person think trace inspired by \citet{bloom1956taxonomy}. These generated traces are used as supervision for training \textsc{INSIDE}.

\FloatBarrier

\begin{center}

\begin{tcolorbox}[
  breakable,
  enhanced,
  colback=gray!5,
  colframe=violet,
  boxrule=0.5pt,
  arc=2pt,
  width=\textwidth,
  title=System Prompt,
  fonttitle=\bfseries\small
]
\begin{lstlisting}[
  basicstyle=\ttfamily\scriptsize,
  breaklines=true,
  breakatwhitespace=true,
  breakindent=0pt,
  columns=fullflexible,
  keepspaces=true,
  showstringspaces=false,
  xleftmargin=0pt,
  framexleftmargin=0pt,
  resetmargins=true
]
You are analyzing a novice student interacting with an AI programming tutor and reconstructing the internal dialogue behind the code edits they make.

This is a retrospective reconstruction task. You are given:
- the full past interaction history, including previously inferred student dialogue
- the most recent submission and the feedback on that submission
- the subsequent submission 

Students differ in background, experience, confidence, and problem-solving style. They may misunderstand feedback, partially apply it, over-apply it, or ignore parts of it. They often make small local edits instead of fully restructuring their code.

Your task has two steps.

### STEP 1: Infer the student's internal state from the context.

Write these fields from a third-person analyst perspective describing the student, not in the student's own voice.

The states correspond to three learning dimensions commonly used in educational theory: cognitive (knowledge and reasoning), affective (emotions and attitudes toward learning), and action (the concrete problem-solving step the student is taking).

Use exactly these tags:

<cognitive>
In one sentence, describe the student's current understanding of the problem or feedback. This may include recalling relevant concepts, interpreting the feedback, applying a rule, analyzing the cause of a bug, or forming a hypothesis about how to fix the code. Include both correct and incorrect beliefs or misconceptions suggested by the code changes.
</cognitive>

<affective>
In one sentence, describe the student's emotional or motivational state toward the task, such as confusion, uncertainty, frustration, confidence, curiosity, or persistence. This reflects how the student is reacting to the feedback and the difficulty of the problem.
</affective>

<action>
In one sentence, describe the concrete programming step the student appears to take next based on the subsequent submission. This should reflect the student's problem-solving action.
</action>

### STEP 2: Generate the student's internal dialogue.

<think>
Write the student's internal dialogue in first-person voice, based on the inferred states above (cognitive, affective, action). 

The dialogue should reflect what the student likely thought before producing the subsequent submission (i.e., the "voice in their head"). It should read like a brief think-aloud explanation of their reasoning.

The thought should sound like a novice programmer thinking through the problem: informal, tentative, and sometimes incomplete. It may include uncertainty, guesses, or small misunderstandings, and may occasionally include questions the student asks themselves.
</think>

Output exactly in this format and nothing else:

<cognitive>...</cognitive>
<affective>...</affective>
<action>...</action>
<think>...</think>
\end{lstlisting}
\end{tcolorbox}

\vspace{0.5em}

\begin{tcolorbox}[
  breakable,
  enhanced,
  colback=gray!5,
  colframe=violet,
  boxrule=0.5pt,
  arc=2pt,
  width=\textwidth,
  title=User Prompt,
  fonttitle=\bfseries\small
]
\begin{lstlisting}[
  basicstyle=\ttfamily\scriptsize,
  breaklines=true,
  breakatwhitespace=true,
  breakindent=0pt,
  columns=fullflexible,
  keepspaces=true,
  showstringspaces=false,
  xleftmargin=0pt,
  framexleftmargin=0pt,
  resetmargins=true
]
Here is the context for the problem they are working on:

[Problem instructions]
{inp["instructions"]}

[Skeleton code]

Fixed:
{inp["skeleton_code_fixed"]}

Todo:
{inp["skeleton_code_todo"]}

==============================
PAST INTERACTION HISTORY
==============================

{past_history}

==============================
CURRENT STEP
==============================

[Student submission]
{current_prev_sub}

[Feedback the student just received]
{current_fb}

[Student's subsequent submission]
{output}
\end{lstlisting}
\end{tcolorbox}

\captionsetup{hypcap=false}

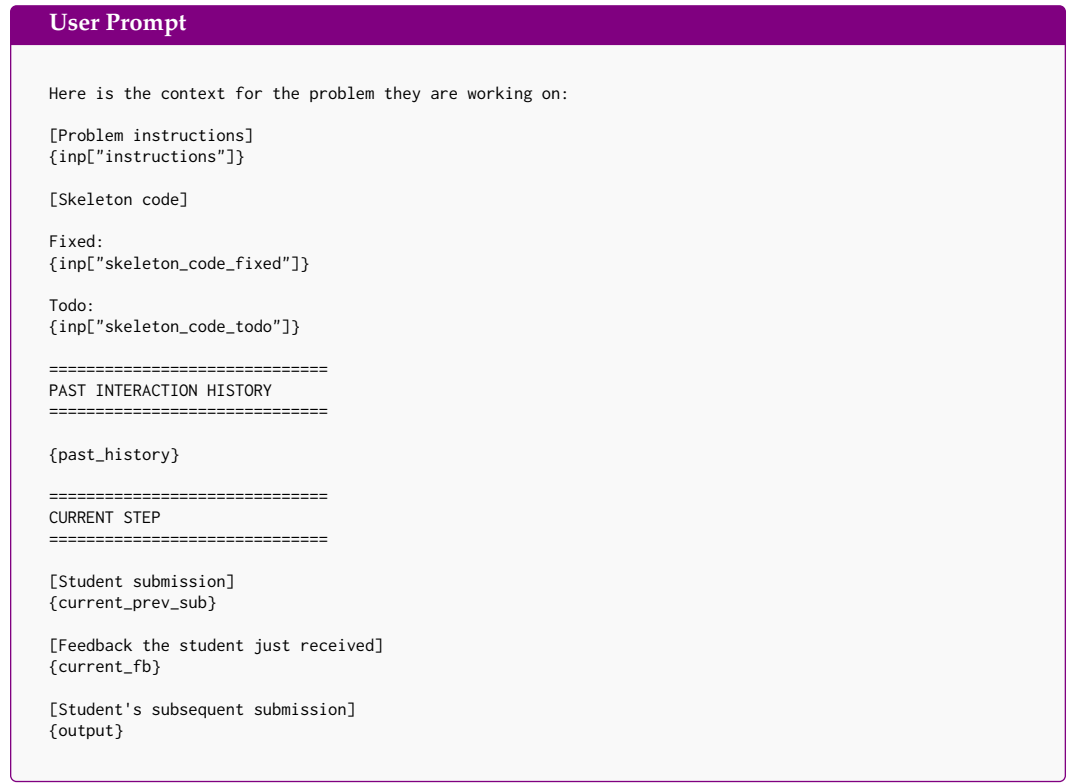
\captionof{figure}{\textbf{Prompt used for internal dialogue generation from real student code submission traces to use for the fine-tuning of INSIDE.}}
\label{fig:dialogue-prompt}
\captionsetup{hypcap=true}

\end{center}

\FloatBarrier

\clearpage
\section{Student Code Simulation Prompts}
\label{app:inference_prompt}

We provide the prompt templates used across all settings: fine-tuning (\autoref{app:sft-prompt}), standard prompting without CoT (\autoref{app:base-prompt}), CoT prompting (\autoref{app:cot-prompt}), and Bloom’s Taxonomy-inspired structured CoT prompting (\autoref{app:bloom-prompt}).

\FloatBarrier

\begin{center}

\begin{tcolorbox}[
  breakable,
  enhanced,
  colback=gray!5,
  colframe=blue,
  boxrule=0.5pt,
  arc=2pt,
  width=\textwidth
]
\begin{lstlisting}[basicstyle=\ttfamily\scriptsize, breaklines=true]
PROBLEM INSTRUCTIONS:
{instructions}

FIXED CODE:
<code>{fixed_code}</code>

TODO CODE:
<code>{skeleton_code}</code>

SUBMISSION HISTORY (CODE + FEEDBACK):
{submissions_with_feedback}
\end{lstlisting}
\end{tcolorbox}

\captionsetup{hypcap=false}
\captionof{figure}{\textbf{Prompt template for fine-tuned models.} The same template is used during both training and inference. Given this input, SFT outputs only the next code submission (\texttt{<code>...</code>}), while \textsc{INSIDE} generates both internal dialogue (\texttt{<think>...</think>}) and the next code (\texttt{<code>...</code>}).}
\label{app:sft-prompt}
\captionsetup{hypcap=true}

\end{center}

\FloatBarrier 
\FloatBarrier

\begin{center}

\begin{tcolorbox}[
  breakable,
  enhanced,
  colback=gray!5,
  colframe=blue,
  boxrule=0.5pt,
  arc=2pt,
  width=\textwidth,
  title=System Prompt,
  fonttitle=\bfseries\small
]
\begin{lstlisting}[
  basicstyle=\ttfamily\scriptsize,
  breaklines=true,
  breakatwhitespace=true,
  breakindent=0pt,
  columns=fullflexible,
  keepspaces=true,
  showstringspaces=false,
  xleftmargin=0pt,
  framexleftmargin=0pt,
  resetmargins=true
]
You are simulating a student taking an introduction to Python programming course.
Generate the student's next attempt at the TODO CODE of the current problem, using both their prior submissions and the feedback they received.

- When writing the code, you must wrap the generated code in <code> and </code> tags.
- Only write the contents of the TODO CODE. Do not include docstrings or anything outside it. 
- As a novice student, your attempts may include mistakes or partial fixes. 
- FIXED CODE is provided - do not copy, modify, or include it.
- Use prior feedback to inform your code revision.
- If the generated attempt is the student's final submission, append "<SUBMIT>" at the end.
- Use the Python programming language
\end{lstlisting}
\end{tcolorbox}

\vspace{0.5em}

\begin{tcolorbox}[
  breakable,
  enhanced,
  colback=gray!5,
  colframe=blue,
  boxrule=0.5pt,
  arc=2pt,
  width=\textwidth,
  title=User Prompt,
  fonttitle=\bfseries\small
]
\begin{lstlisting}[
  basicstyle=\ttfamily\scriptsize,
  breaklines=true,
  breakatwhitespace=true,
  breakindent=0pt,
  columns=fullflexible,
  keepspaces=true,
  showstringspaces=false,
  xleftmargin=0pt,
  framexleftmargin=0pt,
  resetmargins=true
]
PROBLEM INSTRUCTIONS:
{instructions}

FIXED CODE:
<code>{fixed_code}</code>

TODO CODE:
<code>{skeleton_code}</code>

SUBMISSION HISTORY (CODE + FEEDBACK):
{submissions_with_feedback}
\end{lstlisting}
\end{tcolorbox}

\captionsetup{hypcap=false}
\captionof{figure}{\textbf{Prompt template for prompting models without CoT.} The model is instructed to generate only the next code submission (\texttt{<code>...</code>}). These prompts are adapted from prior work \citep{student_code_gen2}, which used LLMs to mimic the distribution of error types and test case failure frequencies of real student datasets.}
\label{app:base-prompt}
\captionsetup{hypcap=true}

\end{center}

\FloatBarrier
\FloatBarrier

\begin{center}

\begin{tcolorbox}[
  breakable,
  enhanced,
  colback=gray!5,
  colframe=blue,
  boxrule=0.5pt,
  arc=2pt,
  width=\textwidth,
  title=System Prompt,
  fonttitle=\bfseries\small
]
\begin{lstlisting}[
  basicstyle=\ttfamily\scriptsize,
  breaklines=true,
  breakatwhitespace=true,
  breakindent=0pt,
  columns=fullflexible,
  keepspaces=true,
  showstringspaces=false,
  xleftmargin=0pt,
  framexleftmargin=0pt,
  resetmargins=true
]
You are simulating a student taking an introduction to Python programming course.
Generate the student's next attempt at the TODO CODE of the current problem, using both their prior submissions and the feedback they received.

Your task has two steps.

### STEP 1: Generate the student's internal dialogue before their next submission using <think> and </think> tags.

<think>
Write the student's internal dialogue in first-person voice. 
The dialogue should reflect what the student likely thought before producing the subsequent submission (i.e., the "voice in their head"). It should read like a brief think-aloud explanation of their reasoning.
The thought should sound like a novice programmer thinking through the problem: informal, tentative, and sometimes incomplete. It may include uncertainty, guesses, or small misunderstandings, and may occasionally include questions the student asks themselves.
</think>

### STEP 2: Generate the student's next code submission.
- Use the internal dialogue above to inform the code the student writes next.
- When writing the code, you must wrap the generated code in <code> and </code> tags.
- Only write the contents of the TODO CODE. Do not include docstrings or anything outside it.
- As a novice student, your attempts may include mistakes or partial fixes.
- FIXED CODE is provided - do not copy, modify, or include it.
- Use prior feedback to inform your code revision.
- If the generated attempt is the student's final submission, append "<SUBMIT>" at the end.
- Use Python programming language
\end{lstlisting}
\end{tcolorbox}

\vspace{0.5em}

\begin{tcolorbox}[
  breakable,
  enhanced,
  colback=gray!5,
  colframe=blue,
  boxrule=0.5pt,
  arc=2pt,
  width=\textwidth,
  title=User Prompt,
  fonttitle=\bfseries\small
]
\begin{lstlisting}[
  basicstyle=\ttfamily\scriptsize,
  breaklines=true,
  breakatwhitespace=true,
  breakindent=0pt,
  columns=fullflexible,
  keepspaces=true,
  showstringspaces=false,
  xleftmargin=0pt,
  framexleftmargin=0pt,
  resetmargins=true
]
PROBLEM INSTRUCTIONS:
{instructions}

FIXED CODE:
<code>{fixed_code}</code>

TODO CODE:
<code>{skeleton_code}</code>

SUBMISSION HISTORY (CODE + FEEDBACK):
{submissions_with_feedback}
\end{lstlisting}
\end{tcolorbox}

\captionsetup{hypcap=false}
\captionof{figure}{\textbf{Prompt template for CoT-based prompting models.} The model is instructed to first generate an internal dialogue (\texttt{<think>...</think>}) and then produce the next code submission (\texttt{<code>...</code>}).}
\label{app:cot-prompt}
\captionsetup{hypcap=true}

\end{center}

\FloatBarrier
\FloatBarrier

\begin{center}

\begin{tcolorbox}[
  breakable,
  enhanced,
  colback=gray!5,
  colframe=blue,
  boxrule=0.5pt,
  arc=2pt,
  width=\textwidth,
  title=System Prompt,
  fonttitle=\bfseries\small
]
\begin{lstlisting}[
  basicstyle=\ttfamily\scriptsize,
  breaklines=true,
  breakatwhitespace=true,
  breakindent=0pt,
  columns=fullflexible,
  keepspaces=true,
  showstringspaces=false,
  xleftmargin=0pt,
  framexleftmargin=0pt,
  resetmargins=true
]
You are simulating a student taking an introduction to Python programming course.
Generate the student's next attempt at the TODO CODE of the current problem, using both their prior submissions and the feedback they received.

Your task has three steps.

## STEP 1: Infer the student's internal state right after receiving the most recent feedback.

Write these fields from a third-person analyst perspective describing the student, not in the student's own voice.
The states correspond to three learning dimensions commonly used in educational theory: cognitive (knowledge and reasoning), affective (emotions and attitudes toward learning), and action (the concrete problem-solving step the student is about to take).

Use exactly these tags:

<cognitive>
In one sentence, describe the student's current understanding of the problem or feedback. This may include recalling relevant concepts, interpreting the feedback, applying a rule, analyzing the cause of a bug, or forming a hypothesis about how to fix the code. Include both correct and incorrect beliefs or misconceptions suggested by the code changes.
</cognitive>

<affective>
In one sentence, describe the student's emotional or motivational state toward the task, such as confusion, uncertainty, frustration, confidence, curiosity, or persistence. This reflects how the student is reacting to the feedback and the difficulty of the problem.
</affective>

<action>
In one sentence, describe the concrete programming step the student is about to take next. This should reflect the student's problem-solving action.
</action>

## STEP 2: Generate the student's internal dialogue before their next submission using <think> and </think> tags.

<think>
Write the student's internal dialogue in first-person voice, based on the inferred states above (cognitive, affective, action). 
The dialogue should reflect what the student likely thought before producing the subsequent submission (i.e., the "voice in their head"). It should read like a brief think-aloud explanation of their reasoning.
The thought should sound like a novice programmer thinking through the problem: informal, tentative, and sometimes incomplete. It may include uncertainty, guesses, or small misunderstandings, and may occasionally include questions the student asks themselves.
</think>

## STEP 3: Generate the student's next code submission.
- Use the internal dialogue above to inform the code the student writes next.
- When writing the code, you must wrap the generated code in <code> and </code> tags.
- Only write the contents of the TODO CODE. Do not include docstrings or anything outside it.
- As a novice student, your attempts may include mistakes or partial fixes.
- FIXED CODE is provided - do not copy, modify, or include it.
- Use prior feedback to inform your code revision.
- If the generated attempt is the student's final submission, append "<SUBMIT>" at the end.
- Use Python programming language
\end{lstlisting}
\end{tcolorbox}

\vspace{0.5em}

\begin{tcolorbox}[
  breakable,
  enhanced,
  colback=gray!5,
  colframe=blue,
  boxrule=0.5pt,
  arc=2pt,
  width=\textwidth,
  title=User Prompt,
  fonttitle=\bfseries\small
]
\begin{lstlisting}[
  basicstyle=\ttfamily\scriptsize,
  breaklines=true,
  breakatwhitespace=true,
  breakindent=0pt,
  columns=fullflexible,
  keepspaces=true,
  showstringspaces=false,
  xleftmargin=0pt,
  framexleftmargin=0pt,
  resetmargins=true
]
PROBLEM INSTRUCTIONS:
{instructions}

FIXED CODE:
<code>{fixed_code}</code>

TODO CODE:
<code>{skeleton_code}</code>

SUBMISSION HISTORY (CODE + FEEDBACK):
{submissions_with_feedback}
\end{lstlisting}
\end{tcolorbox}

\captionsetup{hypcap=false}
\captionof{figure}{\textbf{Bloom's Taxonomy-inspired prompt template for structured CoT prompting.} The model first infers the student's internal state across cognitive, affective, and action dimensions, then generates an internal dialogue (\texttt{<think>...</think>}), followed by the next code submission (\texttt{<code>...</code>}).}
\label{app:bloom-prompt}
\captionsetup{hypcap=true}

\end{center}

\FloatBarrier

\newpage
\section{LLM Judge Evaluation Prompt}
\label{app:eval-prompt}

We provide the full prompt used for LLM-based evaluation of the model-generated internal dialogue alignment in \autoref{fig:eval-prompt}. The prompt instructs the judge (GPT-5-mini) to decompose the generated reasoning into atomic claims and assess whether each claim is supported by the real student's code edits.
\FloatBarrier
\begin{center}
\begin{tcolorbox}[
  breakable,
  enhanced,
  colback=gray!5,
  colframe=orange,
  boxrule=0.5pt,
  arc=2pt,
  width=\textwidth,
  title=System Prompt,
  fonttitle=\bfseries\small
]
\begin{lstlisting}[
  basicstyle=\ttfamily\scriptsize,
  breaklines=true,
  breakatwhitespace=true,
  breakindent=0pt,
  columns=fullflexible,
  keepspaces=true,
  showstringspaces=false,
  xleftmargin=0pt,
  framexleftmargin=0pt,
  resetmargins=true
]
You are evaluating a model that simulates a novice student working on a Python programming assignment.
Extract each distinct claim or intention from the Synthetic Think (e.g., "I will add a base case", "I'll use n % 10"). For each claim, evaluate it against both diffs:
- **Task 1** - Does the claim appear in the **synthetic code diff** (t-1 -> synthetic t)?
- **Task 2** - Does the same claim appear in the **GT code diff** (t-1 -> GT t)?
If a claim is reflected in both, it suggests the synthetic think captures the same intent as the ground truth.

## Output Format
Respond in this exact JSON format:
{
  "claims": [
    {
      "claim": "<claim extracted from synthetic think>",
      "task1_rationale": "<why this claim is/isn't reflected in the synthetic diff>",
      "task1_reflected": true or false,
      "task2_rationale": "<why this claim is/isn't reflected in the GT diff>",
      "task2_reflected": true or false
    },
    {
      "claim": "<claim extracted from synthetic think>",
      "task1_rationale": "<why this claim is/isn't reflected in the synthetic diff>",
      "task1_reflected": true or false,
      "task2_rationale": "<why this claim is/isn't reflected in the GT diff>",
      "task2_reflected": true or false
    }
  ]
}
\end{lstlisting}
\end{tcolorbox}
\vspace{0.5em}
\begin{tcolorbox}[
  breakable,
  enhanced,
  colback=gray!5,
  colframe=orange,
  boxrule=0.5pt,
  arc=2pt,
  width=\textwidth,
  title=User Prompt,
  fonttitle=\bfseries\small
]
\begin{lstlisting}[
  basicstyle=\ttfamily\scriptsize,
  breaklines=true,
  breakatwhitespace=true,
  breakindent=0pt,
  columns=fullflexible,
  keepspaces=true,
  showstringspaces=false,
  xleftmargin=0pt,
  framexleftmargin=0pt,
  resetmargins=true
]
### Problem Description:
{instructions}

### Student's Code at Time t-1 (most recent prior submission):
{code_tm1}

### Feedback the Student Received at Time t-1:
{feedback_tm1}

### Ground Truth Code at Time t (what the real student submitted next):
{gt_code_t}

### Synthetic Think at Time t-1 (what the model thinks the student thought):
{syn_think_tm1}

### Synthetic Code at Time t (what the model predicts the student will submit next):
{syn_code_t}

### Code Diff (t-1 -> synthetic t):
{syn_diff}

### Code Diff (t-1 -> GT t):
{gt_diff}
\end{lstlisting}
\end{tcolorbox}
\captionsetup{hypcap=false}
\captionof{figure}{\textbf{Prompt used for claim-level evaluation of the internal dialogue}. A single call produces both judgments: \texttt{task2\_reflected} against the ground-truth diff gives the alignment score reported in \autoref{sec:eval_internal_dialogue}, and \texttt{task1\_reflected} against the model-generated diff gives the self-consistency score reported in \autoref{app:self-consistency}.}
\label{fig:eval-prompt}
\captionsetup{hypcap=true}
\end{center}
\FloatBarrier

\section{Qualitative Examples}
\label{app:examples}
\subsection{\textsc{INSIDE} vs. Prompting Model}

We provide qualitative examples comparing generated internal dialogue and code against real student behavior in \autoref{fig:comparison}. These examples illustrate how INSIDE captures reasoning that aligns with observed code edits, in contrast to prompting-based methods.

\FloatBarrier
\begin{center}
\begin{tcolorbox}[
  breakable,
  enhanced,
  colback=gray!5,
  colframe=black,
  boxrule=0.5pt,
  arc=2pt,
  width=\textwidth
]
\textbf{Real student code $t$-1:}
\begin{lstlisting}[
  basicstyle=\ttfamily\scriptsize,
  breaklines=true,
  breakatwhitespace=true,
  breakindent=0pt,
  columns=fullflexible,
  keepspaces=true,
  showstringspaces=false,
  xleftmargin=0pt,
  framexleftmargin=0pt,
  resetmargins=true,
  aboveskip=2pt,
  belowskip=2pt,
  escapechar=@
]
def make_repeater(f, n):
    def func(x):
        for _ in range(n):
            f = f(f)
        return f
    return func@\colorbox{red!30}{()}@
    # new = f
    # for _ in range(n):
    #     new = f(new)
    # return new()
\end{lstlisting}

\vspace{0.5em}
\textbf{Feedback $t$-1:}\\
{\ttfamily\scriptsize
You're on the right track with the helper function, but there's a small mistake. In the
loop, you're trying to reassign `f' to `f(f)', but `f' is a function and you should be
applying `f' to an argument, not to the function `f' itself. Also, you're trying to call
`func()' at the end, but `func' should be returned as a function, not called. Try to
apply `f' to `x' in the loop and return `func' instead of `func()'.
}

\vspace{0.5em}
\textbf{Qwen3-8B-BloomCoT Internal dialogue (Alignment: 50\%):}\\
{\ttfamily\scriptsize
Okay, I need to make a function that applies `f' `n' times. I think I should create a
helper function inside `make\_repeater' that takes `x' and applies `f' repeatedly. Let
me try to loop `n' times and apply `f' to the result each time. Oh right, I should start
with `x' and then apply `f' to it in each iteration. Let me make sure I'm not calling
`f' incorrectly.
}

\vspace{0.5em}
\textbf{Qwen3-8B-INSIDE Internal dialogue (Alignment: 100\%):}\\
{\ttfamily\scriptsize
Okay, they said I should apply f to an argument, so maybe I can do f = f(f(x)) inside
the loop and then return f at the end. I'll keep the helper and return it---hopefully
this counts as applying f n times.
}

\vspace{0.5em}
\textbf{Real student code $t$:}
\begin{lstlisting}[
  basicstyle=\ttfamily\scriptsize,
  breaklines=true,
  breakatwhitespace=true,
  breakindent=0pt,
  columns=fullflexible,
  keepspaces=true,
  showstringspaces=false,
  xleftmargin=0pt,
  framexleftmargin=0pt,
  resetmargins=true,
  aboveskip=2pt,
  belowskip=2pt,
  escapechar=@
]
def make_repeater(f, n):
    def func(x):
        for _ in range(n):
            f = f(f@\colorbox{green!30}{(x)}@)
        return f
    return func
    # new = f
    # for _ in range(n):
    #     new = f(new)
    # return new()
\end{lstlisting}
\end{tcolorbox}
\captionsetup{hypcap=false}
\captionof{figure}{\textbf{A qualitative comparison of synthetic internal dialogue generated
by Qwen3-8B-BloomCoT and Qwen3-8B-INSIDE for the same student code and feedback}
(\textcolor{red}{red highlights} indicate code removed from the previous student
submission, while \textcolor{green!60!black}{green highlights} indicate code added
in the next student submission). BloomCoT generates a clean, competent think trace
reasoning toward the correct solution, yet the resulting student code remains buggy and
misaligned with that reasoning. INSIDE generates a messier, uncertain think trace that
mirrors how a novice might partially misinterpret the feedback --- producing
\texttt{f = f(f(x))} as a guess --- which faithfully matches the actual student code
change. This example illustrates that INSIDE captures the kind of confused,
partially-correct reasoning real students exhibit, rather than generating idealized
reasoning decoupled from the student's actual behavior.}
\label{fig:comparison}
\captionsetup{hypcap=true}
\end{center}
\FloatBarrier

\newpage
\subsection{Failure Mode Analysis}
\label{app:failure-modes}

We conducted a qualitative error analysis of \textsc{INSIDE}'s failure cases. From the generated
traces of the \textsc{INSIDE} models, we randomly sampled 25 examples from the bottom quartile
of alignment scores and inspected the student's code $c_{t-1,s_i,p_u}$, the feedback
$f_{t-1,s_i,p_u}$, the generated internal dialogue $\hat{z}_{t,s_i,p_u}$, and the
student's next submission $c_{t,s_i,p_u}$. We grouped the observations into three
recurring failure categories: \textbf{no student edit} (48\%), where the student
resubmitted the same code despite the feedback; \textbf{overestimating feedback uptake}
(44\%), where the generated dialogue described a more feedback-responsive repair than
the student implemented; and \textbf{underestimating feedback uptake} (8\%), where the
dialogue failed to anticipate a partial improvement the student made. Representative
examples are shown in \autoref{fig:failure-no-edit}, \autoref{fig:failure-overestimate},
and \autoref{fig:failure-underestimate}.

\FloatBarrier
\begin{center}
\begin{tcolorbox}[
  breakable,
  enhanced,
  colback=gray!5,
  colframe=black,
  boxrule=0.5pt,
  arc=2pt,
  width=\textwidth
]
\textbf{Real student code $t$-1:}
\begin{lstlisting}[
  basicstyle=\ttfamily\scriptsize,
  breaklines=true,
  breakatwhitespace=true,
  breakindent=0pt,
  columns=fullflexible,
  keepspaces=true,
  showstringspaces=false,
  xleftmargin=0pt,
  framexleftmargin=0pt,
  resetmargins=true,
  aboveskip=2pt,
  belowskip=2pt,
  escapechar=@
]
def num_eights(n):
\end{lstlisting}
\vspace{0.5em}
\textbf{Feedback $t$-1:}\\
{\ttfamily\scriptsize
It seems like you haven't written any code for the num\_eights function yet.
Remember, you need to use recursion to solve this problem. Think about how you can
break down the problem: you can check if the last digit of the number is 8, and then
recursively call the function on the rest of the number. How can you get the last
digit and the rest of the number?
}
\vspace{0.5em}

\textbf{Qwen3-8B-INSIDE Internal dialogue:}\\
{\ttfamily\scriptsize
Okay, I can check the last digit with n \% 10 and if it's 8 add 1, then call
the function on n // 10. I'll write a base case for n == 0 and just
return 0 there---this should work without loops or assignments.
}
\vspace{0.5em}

\textbf{Real student code $t$:}
\begin{lstlisting}[
  basicstyle=\ttfamily\scriptsize,
  breaklines=true,
  breakatwhitespace=true,
  breakindent=0pt,
  columns=fullflexible,
  keepspaces=true,
  showstringspaces=false,
  xleftmargin=0pt,
  framexleftmargin=0pt,
  resetmargins=true,
  aboveskip=2pt,
  belowskip=2pt,
  escapechar=@
]
def num_eights(n):
\end{lstlisting}
\end{tcolorbox}
\captionsetup{hypcap=false}
\captionof{figure}{\textbf{Representative example for ``No student edit.''} Despite receiving
detailed feedback outlining a recursive approach, the real student resubmits an empty
function stub unchanged. The INSIDE-generated internal dialogue, in contrast, reflects
a plausible and largely correct plan for solving the problem (using \texttt{n \% 10}
and \texttt{n // 10} with a base case), illustrating a case where the model's inferred
reasoning is more feedback-responsive than the student's actual (lack of) action.}
\label{fig:failure-no-edit}
\captionsetup{hypcap=true}
\end{center}
\FloatBarrier
\FloatBarrier
\begin{center}
\begin{tcolorbox}[
  breakable,
  enhanced,
  colback=gray!5,
  colframe=black,
  boxrule=0.5pt,
  arc=2pt,
  width=\textwidth
]
\textbf{Real student code $t$-1:}
\begin{lstlisting}[
  basicstyle=\ttfamily\scriptsize,
  breaklines=true,
  breakatwhitespace=true,
  breakindent=0pt,
  columns=fullflexible,
  keepspaces=true,
  showstringspaces=false,
  xleftmargin=0pt,
  framexleftmargin=0pt,
  resetmargins=true,
  aboveskip=2pt,
  belowskip=2pt,
  escapechar=@
]
def num_eights(n):
    if n // 10 == 0 and n % 10 == 8:
        return 1
    elif n // 10 == 0 and n % 10 != 8:
        return 0
    else:
        return num_eights(n // 10) @\colorbox{red!30}{+ 1}@
\end{lstlisting}

\vspace{0.5em}
\textbf{Feedback $t$-1:}\\
{\ttfamily\scriptsize
You're almost there! However, you're currently adding 1 to the count of 8's in every
recursive call, not just when the last digit is 8. You should only add 1 if the last
digit is 8. Try to fix this issue and run your code again.
}

\vspace{0.5em}
\textbf{Qwen3-8B-INSIDE Internal dialogue:}\\
{\ttfamily\scriptsize
Oh, I was adding 1 every time---so I should only add 1 when the last digit is 8. I'll
keep my base cases and just return 1 + num\_eights(n // 10) when n \% 10 == 8, otherwise
just recurse on n // 10.
}

\vspace{0.5em}
\textbf{Real student code $t$:}
\begin{lstlisting}[
  basicstyle=\ttfamily\scriptsize,
  breaklines=true,
  breakatwhitespace=true,
  breakindent=0pt,
  columns=fullflexible,
  keepspaces=true,
  showstringspaces=false,
  xleftmargin=0pt,
  framexleftmargin=0pt,
  resetmargins=true,
  aboveskip=2pt,
  belowskip=2pt,
  escapechar=@
]
def num_eights(n):
    if n // 10 == 0 and n % 10 == 8:
        return 1
    elif n // 10 == 0 and n % 10 != 8:
        return 0
    else:
        return@\colorbox{green!30}{(1 + }@num_eights(n // 10)@\colorbox{green!30}{)}@
\end{lstlisting}
\end{tcolorbox}
\captionsetup{hypcap=false}
\captionof{figure}{\textbf{Representative example for ``Overestimating feedback uptake''}
(\textcolor{red}{red highlights} indicate code removed from the previous student
submission, while \textcolor{green!60!black}{green highlights} indicate code added
in the next student submission). The generated internal dialogue anticipates a targeted
fix conditioning the increment on \texttt{n \% 10 == 8}, but the real student's actual
edit only reorders the existing expression, moving \texttt{+ 1} in front of the recursive
call and wrapping it in parentheses.}
\label{fig:failure-overestimate}
\captionsetup{hypcap=true}
\end{center}
\FloatBarrier
\FloatBarrier
\begin{center}
\begin{tcolorbox}[
  breakable,
  enhanced,
  colback=gray!5,
  colframe=black,
  boxrule=0.5pt,
  arc=2pt,
  width=\textwidth
]
\textbf{Real student code $t$-1:}
\begin{lstlisting}[
  basicstyle=\ttfamily\scriptsize,
  breaklines=true,
  breakatwhitespace=true,
  breakindent=0pt,
  columns=fullflexible,
  keepspaces=true,
  showstringspaces=false,
  xleftmargin=0pt,
  framexleftmargin=0pt,
  resetmargins=true,
  aboveskip=2pt,
  belowskip=2pt,
  escapechar=@
]
def make_repeater(f, n):
    k = 0
    x = ()
    while k < n:
        @\colorbox{red!30}{x}@ = f(x)
        k += 1
    return f(x)
\end{lstlisting}

\vspace{0.5em}
\textbf{Feedback $t$-1:}\\
{\ttfamily\scriptsize
You're still returning a value, not a function. Remember, make\_repeater should return a
function that, when called with an argument x, applies f to x n times. This returned
function should take one argument. Also, you're still trying to use x before it's defined.
You should define x inside the function you're returning, not in make\_repeater.
}

\vspace{0.5em}
\textbf{Qwen2.5-Coder-7B-INSIDE Internal dialogue:}\\
{\ttfamily\scriptsize
They keep saying return a function, but maybe if I just loop n times and update x with
f(x), then return x, that shows it applied f multiple times. I'm not sure how to make the
returned function take x, but at least this uses f in a loop and returns something.
}

\vspace{0.5em}
\textbf{Real student code $t$:}
\begin{lstlisting}[
  basicstyle=\ttfamily\scriptsize,
  breaklines=true,
  breakatwhitespace=true,
  breakindent=0pt,
  columns=fullflexible,
  keepspaces=true,
  showstringspaces=false,
  xleftmargin=0pt,
  framexleftmargin=0pt,
  resetmargins=true,
  aboveskip=2pt,
  belowskip=2pt,
  escapechar=@
]
def make_repeater(f, n):
    k = 0
    while k < n:
        @\colorbox{green!30}{f}@ = f(x)
        k += 1
    return @\colorbox{green!30}{lambda x: f(x)}@
\end{lstlisting}
\end{tcolorbox}
\captionsetup{hypcap=false}
\captionof{figure}{\textbf{Representative example for ``Underestimating feedback uptake''}
(\textcolor{red}{red highlights} indicate code removed from the previous student
submission, while \textcolor{green!60!black}{green highlights} indicate code added
in the next student submission). The generated internal dialogue anticipates only an
incremental fix within the existing loop-based structure, but the real student makes a
more substantial revision by introducing a \texttt{lambda} to return a function, a change
the dialogue fails to anticipate.}
\label{fig:failure-underestimate}
\captionsetup{hypcap=true}
\end{center}
\FloatBarrier

\end{document}